\documentclass[smallcondensed]{svjour3}     % onecolumn (ditto)
\usepackage{cite}
\usepackage{amsmath,amssymb,amsfonts}
\usepackage{graphicx}
\usepackage{textcomp}
\usepackage{url}
\usepackage{subfig}
\usepackage[usenames, dvipsnames]{xcolor}
\usepackage{enumitem}

\usepackage{natbib}
\usepackage{soul} % for \st

\usepackage{amsmath}
\usepackage{algorithm}
\usepackage{algpseudocode}
\def\BState{\State\hskip-\ALG@thistlm}
\algrenewcommand\algorithmicrequire{\textbf{Input:}}
\begin{document}

\title{Data Stream Clustering: A Review
}


\author{Alaettin Zubaro\u{g}lu      \and
        Volkan Atalay
}

\institute{A. Zubaro\u{g}lu \at
              Department of Computer Engineering \\
              Middle East Technical University \\
              Dumlup{\i}nar Bulvar{\i} No:1 06800 \c{C}ankaya, Ankara, Turkey \\
              Tel.: +90-505-323-3240 \\
              \email{alaettin.zubaroglu@metu.edu.tr}           %  \\
           \and
           V. Atalay \at
              Department of Computer Engineering \\
              Middle East Technical University \\
              Dumlup{\i}nar Bulvar{\i} No:1 06800 \c{C}ankaya, Ankara, Turkey \\
              Tel.: +90-312-210-5576 \\
              \email{vatalay@metu.edu.tr} 
}

\date{Received: date / Accepted: date}

\maketitle

\begin{abstract}
Number of connected devices is steadily increasing and these devices continuously generate data streams. Real-time processing of data streams is arousing interest despite many challenges. Clustering is one of the most suitable methods for real-time data stream processing, because it can be applied with less prior information about the data and it does not need labeled instances. However, data stream clustering differs from traditional clustering in many aspects and it has several challenging issues. 
Here, we provide information regarding the concepts and common characteristics of data streams, such as concept drift, data structures for data streams, time window models and outlier detection. We comprehensively review recent data stream clustering algorithms and analyze them in terms of the base clustering technique, computational complexity and clustering accuracy. A comparison of these algorithms is given along with still open problems. We indicate popular data stream repositories and datasets, stream processing tools and platforms. Open problems about data stream clustering are also discussed.
\end{abstract}

%%%\begin{IEEEkeywords}
%%%stream clustering, online clustering, data streams
%%%\end{IEEEkeywords}

% For table of contents, enable following line.
%%%\tableofcontents

%
\section{Introduction}
More devices including sensors are becoming interconnected and interconnected devices continuously generate streams of data at high speed. Offline processing of such huge amount of data requires growing storage capacity and may cause delayed analyses. Hence, real-time processing of the data generated by the connected devices has become an active research area. 

A data stream is a potentially unbounded, ordered sequence of instances. A data stream $S$ may be shown as $S = \{\textbf{x}_1, \textbf{x}_2, \textbf{x}_3, ..., \textbf{x}_N\}$ where $\textbf{x}_i$ is $i^{th}$ data instance, which is a $d$-dimensional feature vector and $N$ goes to infinity. Data stream differs from the traditional, stored data in many aspects. In most cases, true class labels are not available for stream instances and there is no prior knowledge about the number of classes. Therefore, clustering, being unsupervised is one of the most suitable data mining and data analysis methods for data streams. Border security using sensors, auto monitoring of surveillance cameras, internet of things (IoT) device tracking, real time patient tracking, stock market analysis, network intrusion detection, and earthquake forecasting systems are among the applications of data stream clustering.

Data clustering, is the task of grouping instances such that the instances in the same group are similar to each other and the instances in different groups are dissimilar according to the properties of the instances. Hence, the objective of clustering is to minimize intra-cluster distance and maximize inter-cluster distance. 
However, data stream clustering differs from traditional clustering in many aspects and it has several challenging issues. Data stream instances can be read only once, in a certain order, and must be processed in a short time interval, before the next instance is received. Data streams can not be stored, only a synopsis of the stream is stored, if required. 
Table~\ref{tbl:streamVsTraditional} gives the comparison of stream clustering with traditional clustering (\cite{Mousavi:2015:review, gaber2009data}).

%table
\begin{table}[htbp]
\caption{Comparison of stream clustering with traditional clustering.}
\begin{center}
\begin{tabular}{l l}
	%\hline
	\textbf{Stream Clustering} 			& \textbf{Traditional Clustering} \\
	\hline
	Real time processing 		& Offline processing \\
	Data arrives on the fly 	& All data are ready \\
	Only single pass on data is possible & Multiple passes are possible \\
	Data are not feasible to be stored	& Data are suitable to be stored \\
	Only synopsis of the data is stored & All raw data are stored \\
	Approximate results are accepted	& Accurate results are expected \\	
	\hline
\end{tabular}
\label{tbl:streamVsTraditional}
\end{center}
\end{table}

Data stream instances may evolve over time, and this is called \textit{concept drift}. Concept drift is the unforeseen change in the properties of the input data instances. For the case of the traditional data, the whole dataset is available and properties of the instances do not change during the processing of the data. This makes the concept drift a data stream specific challenge. A data stream clustering algorithm should detect and adopt concept drift for more accurate results. According to the occurrence style, different types of concept drift exist. 

Similar to traditional data, data streams may include outliers. To achieve better performance, outliers in the data streams should be detected, interpreted and possibly removed. In data streams, it is not easy to mark an instance as outlier, because a dissimilar instance might be the first sample of a new, previously unseen cluster, i.e. it might be a precursor of a concept drift. Moreover, a dissimilar instance might be marker of an anomaly, which is very valuable for anomaly detection systems. 

Data stream clustering algorithms use special data structures to keep synopsis of the input data, since it is not possible to store the whole data. Storing agglomerative sum or storing only representative samples of the data are two popular alternative structures. Moreover, users are often interested in the most recent data instances rather than the previous ones. This situation creates a requirement of obsolescence for previous data instances. In data stream clustering, it is solved by time window models.

Most of the data stream clustering algorithms use a two phase approach~\citep{Silva:2013survey}.
In \textit{online phase} which is also called as data abstraction phase, a synopsis of the data stream is generated and stored in specialized data structures. Synopsis of the data stream is updated when a new instance is received. Therefore the synopsis always remains up-to-date.
\textit{Offline phase}, called also as clustering phase, runs periodically or whenever the user requests. In this phase, the final clustering is performed over the generated data synopsis.
There also exist several fully online data stream clustering algorithms, which re-cluster the data for every new instance and keep an up-to-date clustering result. Among fully online stream clustering algorithms are DPClust~\citep{DPClust}, CEDAS~\citep{Hyde:2017:Cedas}, DBIECM~\citep{Zhang:2017:dbiecm}, FEAC-Stream~\citep{Silva:2017:FEAC-Stream} and Adaptive Stream $k$-means~\citep{Puschmann:2017:adaptiveStreaming}.

For the evaluation of data stream clustering, traditional techniques are still valid and they are commonly used. 
A relatively new concept \textit{edge computing}~\citep{ShiCao:2016:Edge, ShiDustdar:2016:Edge, Satyanarayanan2017edge} is the technique to process the produced data on several edge nodes that are close to the connected devices, instead of a single central system. It is also an interest arousing novel concept, however it is out of scope of this study. We examine central data stream clustering concept that runs on a single center for the whole system. 

In this manuscript, Section~\ref{sec:streamClustering} is devoted to issues in data stream clustering. We give information about some mechanisms of stream clustering, which are data structures, time window models, concept drift and outlier detection methods. In Section~\ref{sec:methods}, we give brief information about the categories of stream clustering algorithms. Moreover, we examine seven most recent data stream clustering algorithms that are not mentioned in the previous surveys in more detail and explain them one by one. We make a comparative review of the examined algorithms and highlight their advantages and disadvantages against each other in Section~\ref{sec:methodComparison}. We summarize the open problems about data stream clustering in Section~\ref{sec:openProblems}. We indicate popular stream data repositories and datasets, stream processing tools and stream processing platforms in Section~\ref{sec:dataReposSets}, Section~\ref{sec:streamTools} and Section~\ref{sec:streamPlatforms} respectively, before concluding the study in Section~\ref{sec:conclusion}.

\section{Concepts in Data Stream Clustering}
\label{sec:streamClustering}

The information given here that are the basic concepts used in data stream clustering facilitates explaining the recent data clustering algorithms analyzed in Section~\ref{sec:methods}.

\subsection{Concept Drift}
\label{subsub:conceptDrift}

Concept drift is the unforeseen change in statistical properties of data stream instances over time. There are four types of concept drift: sudden, gradual, incremental and recurring~\citep{RamirezDataPreprocessing}. 
%Visualization of these four types of drifts is presented in Figure~\ref{fig:typesOfConceptDrift}.

\begin{itemize}
	\item \textit{Sudden concept drift:} Between two consecutive instances, the change occurs at once, and after this time only instances of the new class are received. An instance that has properties of the previous class never arrives again. A data stream containing sudden concept drift might look like as follows, where different colors indicate different classes.
	
	$S = \{..., \textcolor{red}{\textbf{x}_1}, \textcolor{red}{\textbf{x}_2}, \textcolor{red}{\textbf{x}_3}, \textcolor{red}{\textbf{x}_4}, \textcolor{red}{\textbf{x}_5}, \textcolor{red}{\textbf{x}_6}, \textcolor{blue}{\textbf{x}_7}, \textcolor{blue}{\textbf{x}_8}, \textcolor{blue}{\textbf{x}_9}, \textcolor{blue}{\textbf{x}_{10}}, \textcolor{blue}{\textbf{x}_{11}}, \textcolor{blue}{\textbf{x}_{12}}, ... \}$
	
	\item \textit{Gradual concept drift:} The number of instances belonging to the previous class decreases gradually while the number of instances belonging to the new class increases over time. During a gradual concept drift, instances of both previous and new classes are visible. A data stream containing gradual concept drift might look like as follows, where different colors indicate different classes.
	
	$S = \{..., \textcolor{red}{\textbf{x}_1}, \textcolor{red}{\textbf{x}_2}, \textcolor{red}{\textbf{x}_3}, \textcolor{blue}{\textbf{x}_4}, \textcolor{red}{\textbf{x}_5}, \textcolor{red}{\textbf{x}_6}, \textcolor{blue}{\textbf{x}_7}, \textcolor{blue}{\textbf{x}_8}, \textcolor{red}{\textbf{x}_9}, \textcolor{blue}{\textbf{x}_{10}}, \textcolor{blue}{\textbf{x}_{11}}, \textcolor{blue}{\textbf{x}_{12}}, ... \}$
	
	\item \textit{Incremental concept drift:} Data instances belonging to the previous class evolves to a new class step by step. After the concept drift is completed, the previous class disappears. The instances that arrive during the concept drift are of transitional forms and they do not have to belong to either of the classes. A data stream containing incremental concept drift might look like as follows, where different colors indicate different classes.
	
	$S = \{..., \textcolor{red}{\textbf{x}_1}, \textcolor{red}{\textbf{x}_2}, \textcolor{red}{\textbf{x}_3}, \textcolor{magenta}{\textbf{x}_4}, \textcolor{magenta}{\textbf{x}_5}, \textcolor{purple}{\textbf{x}_6}, \textcolor{purple}{\textbf{x}_7}, \textcolor{violet}{\textbf{x}_8}, \textcolor{violet}{\textbf{x}_9}, \textcolor{blue}{\textbf{x}_{10}}, \textcolor{blue}{\textbf{x}_{11}}, \textcolor{blue}{\textbf{x}_{12}}, ... \}$
	
	\item \textit{Recurring concept drift :} The data instances change between two or more statistical characteristics several times. Neither of the classes disappears permanently but both of them arrive in turns. A data stream containing recurring concept drift might look like as follows, where different colors indicate different classes.
	
	$S = \{..., \textcolor{red}{\textbf{x}_1}, \textcolor{red}{\textbf{x}_2}, \textcolor{red}{\textbf{x}_3}, \textcolor{blue}{\textbf{x}_4}, \textcolor{blue}{\textbf{x}_5}, \textcolor{blue}{\textbf{x}_6}, \textcolor{red}{\textbf{x}_7}, \textcolor{red}{\textbf{x}_8}, \textcolor{red}{\textbf{x}_9}, \textcolor{blue}{\textbf{x}_{10}}, \textcolor{blue}{\textbf{x}_{11}}, \textcolor{blue}{\textbf{x}_{12}}, ... \}$
	
\end{itemize}

Creation of new clusters, disappearance or evolution of existing clusters are all examples of concept drift. Concept drift may also affect the cluster boundaries. If the cluster boundaries are modified, it is called \textit{real concept drift} while in the other case, it is called \textit{virtual concept drift}. 
There exist several studies in the literature for concept drift detection. 
%Gama et al. have a comprehensive survey~\cite{conceptDriftSurvey} on concept drift detection. 
\citet{conceptDriftSurvey} have a comprehensive survey on concept drift detection.

\subsection{Data Structures for Data Streams}
\label{subsec:dataStruct}
In data stream processing, it is not possible to store the whole input data because data streams are infinite and all existing processing systems have main memory constraint. Therefore, only a synopsis of the input stream is stored and this situation makes it essential to develop special data structures that enables to incrementally summarize the input stream. Four most commonly used data structures are \textit{feature vectors, prototype arrays, coreset trees \textnormal{and} grids}. Feature vectors keep the summary of the data instances, prototype arrays keep only a number of representative instances that exemplify the data, coreset trees keep the summary in a tree structure and grids keep the data density in the feature space~\citep{Silva:2013survey, Ghesmoune:2016:review, Mansalis:2018:review}.

\subsection{Time Window Models}
\label{subsec:timeWindowModels}
In data stream processing, it is more efficient to process recent data instead of the whole data. Different window models are developed for this purpose. There are three different window models, which are damped window, landmark window and sliding window models. These window models are presented in Figure~\ref{fig:windowModels}.
 
\subsubsection{Damped Window}
\label{subsubsec:dampedwindow}
In damped window model, recent data have more weight than the older data. The most recent instance has the most weight and the importance of the instances decreases by time. This method is usually implemented using decay functions which scale down the weight of the instances, depending on the time passed since the instance is received. One of such functions is $f(t) = 2\textsuperscript{-$\lambda$t}$,
% is given in Equation~\ref{equ:fading}, 
where \textit{t} is the time passed and $\lambda$ is the decay rate. Higher decay rate in the function means a more rapid decrease in the value. Figure~\ref{fig:windowModels} (a) demonstrates the damped window model.

\subsubsection{Landmark Window}
In landmark window model, the whole data between two landmarks are included in the processing and all of the instances have equal weight. Amount of data that belong to a single window is called \textit{window length} and usually indicated by \textit{w}. Window length can be defined as instance count or elapsed time. In landmark window method, consecutive windows do not intersect and the new window just begins from the point the previous window ends. According to this definition, data instances belong to a window are calculated using Equation~\ref{equ:dataInstances} and window number of a data instance is calculated using Equation~\ref{equ:windowNumber} where $w$ is window length, $x_i$ is $i^{th}$ instance and $W_m$ is $m^{th}$ window. Indexes $i$ and $m$ start with zero. Figure~\ref{fig:windowModels} (b) shows the landmark window model.

\begin{equation}
\label{equ:dataInstances}
W_m = [x_{m*w}, ... ,x_{(m+1)*w-1}]
\end{equation}

\begin{equation}
\label{equ:windowNumber}
m = \left \lfloor{ \frac{i}{w} }\right \rfloor
\end{equation}

\subsubsection{Sliding Window}
\begin{figure}%
 	\centering
 	\subfloat[Damped window model.]{{\includegraphics[width = 8cm]{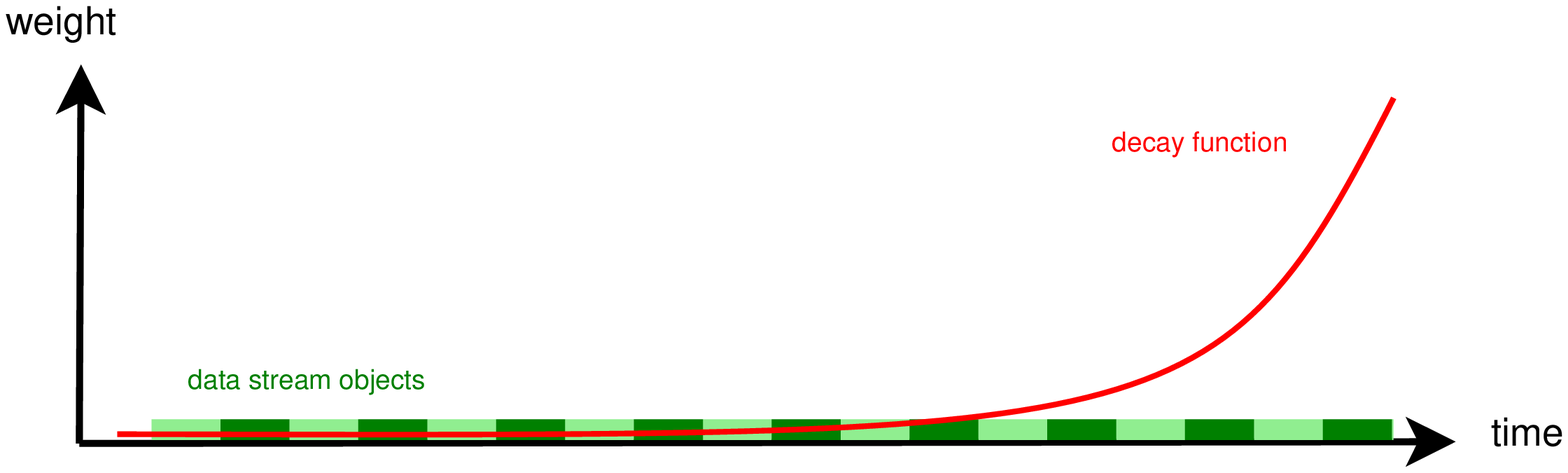} }}%
 	\qquad
 	\vspace{2mm}
 	\subfloat[Landmark window model.]{{\includegraphics[width = 8cm]{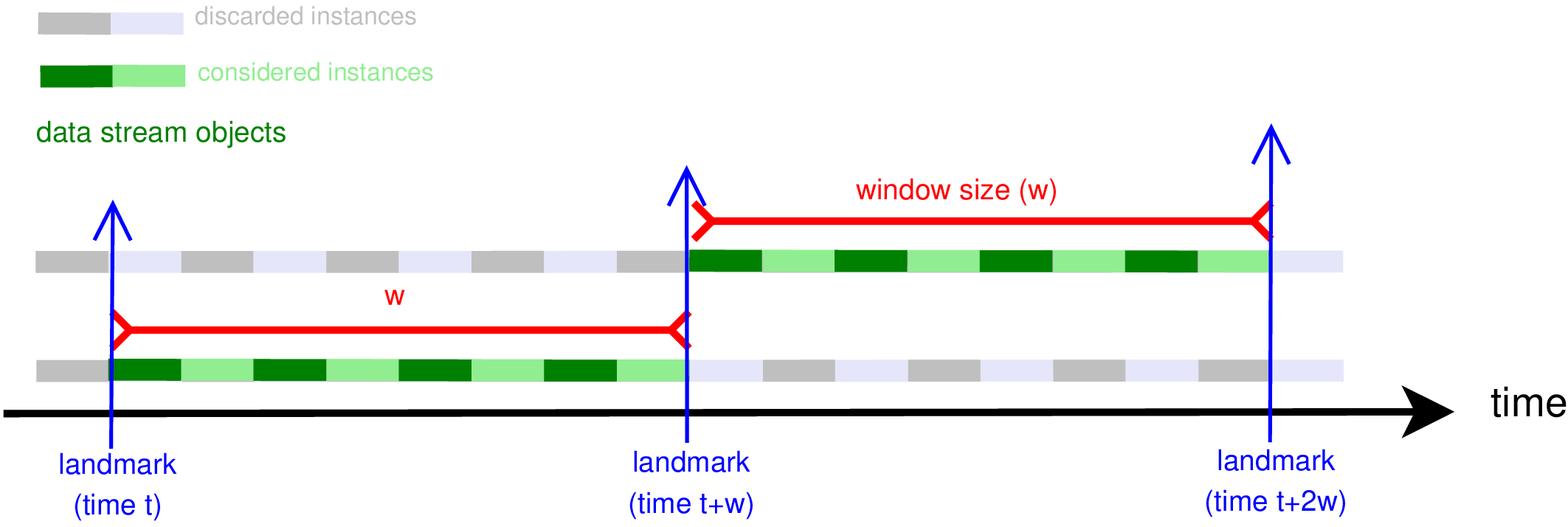} }}%
 	\qquad
 	\vspace{2mm}
 	\subfloat[Sliding window model.]{{\includegraphics[width = 8cm]{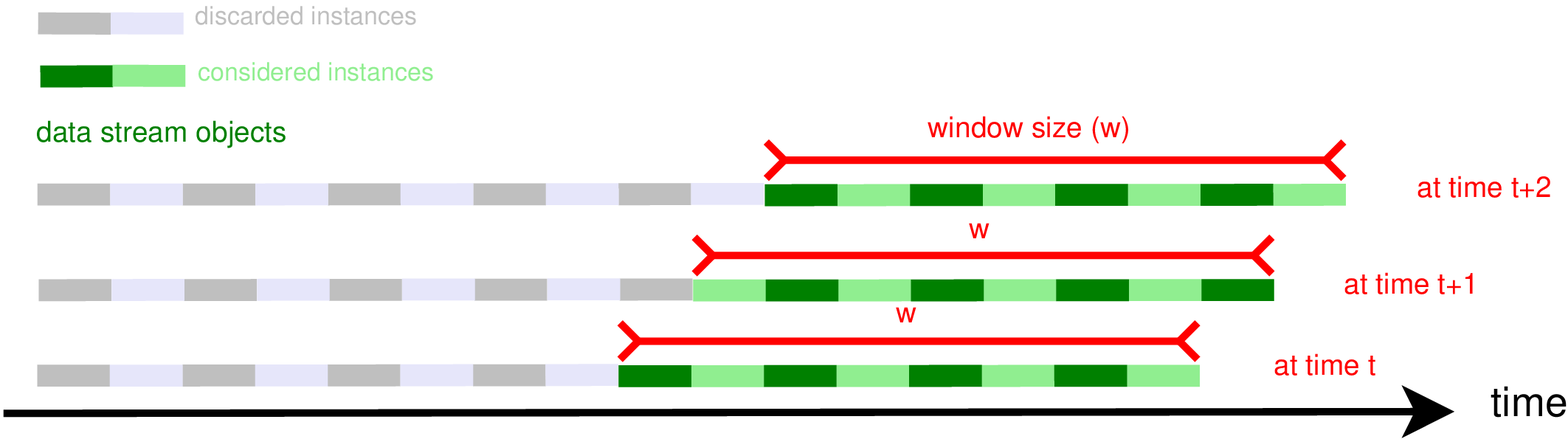} }}%
 	\caption{Time window models.}%
 	\label{fig:windowModels}%
 \end{figure}

In sliding window model, the window swaps one instance at each step. The older instance moves out of the window, and the most recent instance moves in to the window by FIFO style. All instances in the window have equal weight and consecutive windows mostly overlap. Window length is a user defined parameter and should be decided according to the input data. Figure~\ref{fig:windowModels} (c) describes this window model and data instances belong to a window are calculated using Equation~\ref{equ:slidingInstances} where $w$ is window length, $x_i$ is $i^{th}$ instance and $W_m$ is $m^{th}$ window. Indexes $i$ and $m$ start with zero. Figure~\ref{fig:windowModels} (c) presents the sliding window model.

\begin{equation}
\label{equ:slidingInstances}
W_m = [x_{m}, ... ,x_{(m+w-1)}]
\end{equation}
 
\subsection{Outlier Detection}
\label{subsec:outlierDetection}
Outlier is a data instance that seems to be different from the remaining of the data. Either it does not belong to any of the clusters, or it belongs to a cluster whose cardinality is far less than other clusters. Let $C_i$ be $i^{th}$ cluster, $|C_i|$ be cardinality of $C_i$ and $k$ be cluster count, if $|C_j| << \frac {1} {k} {\sum_{i=1}^{k} |C_i|}$, then $C_j$ is treated as outlier. There exist several definitions for the outliers in the literature~\citep{Modi2017OutlierAA}. An outlier can occur because of malicious activities, instrumental errors, transmission problems, data collection problems or similar~\citep{Merino2015}. In data mining, outliers negatively affect the processing accuracy, because of that, outlier detection has a crucial importance in data mining. It is possible to benefit from several existing surveys~\citep{Modi2017OutlierAA, Chauhan2015outlier, Souiden2016outlier, Thakkar2016outlier, Bhosale2014outlier, Sadik:2014outlier} about outlier detection in data streams. 
\citet{Thakkar2016outlier} have classified outlier detection techniques in four main groups in their survey.

\begin{itemize}
	\item \textit{Statistical outlier detection} methods make an assumption about the data distribution. Taking the distribution into account, data instances that have a low probability to be generated, are marked as outliers. Statistical outlier detection methods are divided into two categories: parametric methods and non-parametric methods. In parametric methods, a distribution model of the data is assumed before starting, according to the parameters. This method is not suitable for data streams, since the entire dataset is not available in streams and the distribution model may change over time. In non-parametric methods, no distribution model is assumed a priori; instead, the distribution model is learned from the original data instances. This property makes non-parametric statistical outlier detection methods adoptable for data streams.
	\item \textit{Distance based outlier detection} methods~\citep{Chauhan2015outlier} use neighbor count of the instance to decide whether it is an outlier or not. Two parameters $R$ and $k$ play the main role. If the data instance has less than $k$ neighbors in a distance of $R$, then it is marked as an outlier. A distance measure (or a similarity measure) must be defined. No domain knowledge is required and no distribution model assumption is done. Therefore, distance based outlier detection methods are suitable for data streams. However, they are not effective for high dimensional data streams. 
	\item \textit{Density based outlier detection} methods compare the density around the data instance to the density around its neighbors. If the instance has a density around it similar to its neighbors, then it is not an outlier. Otherwise it is considered as an outlier. This group of methods, are more effective than distance based methods, however they have a higher computational complexity. 
	\item In \textit{Clustering based outlier detection} methods~\citep{Bhosale2014outlier} data instances that do not belong to any clusters, or far away from their cluster centroids, are potential outliers. Moreover, outliers may belong to a sparse or small cluster that is not close to other clusters. Ordinary data instances are expected to belong to large, dense clusters and they are relatively close to cluster centroids.
\end{itemize}

Although real-time analysis of data streams is a more recent research subject, it has many similarities with the analysis of time series data which has been studied for longer time and more abundant in the literature. Especially outlier detection in data streams is very similar to anomaly detection in time series data analysis~\citep{Kong2019timeseries, Christodoulou2018timeseries, Keogh2005timeseries}.
\section{Stream Clustering Algorithms}
\label{sec:methods}

%\subsection{Existing Surveys}

\begin{figure*}[t]
	\centerline{\includegraphics[width = \textwidth]{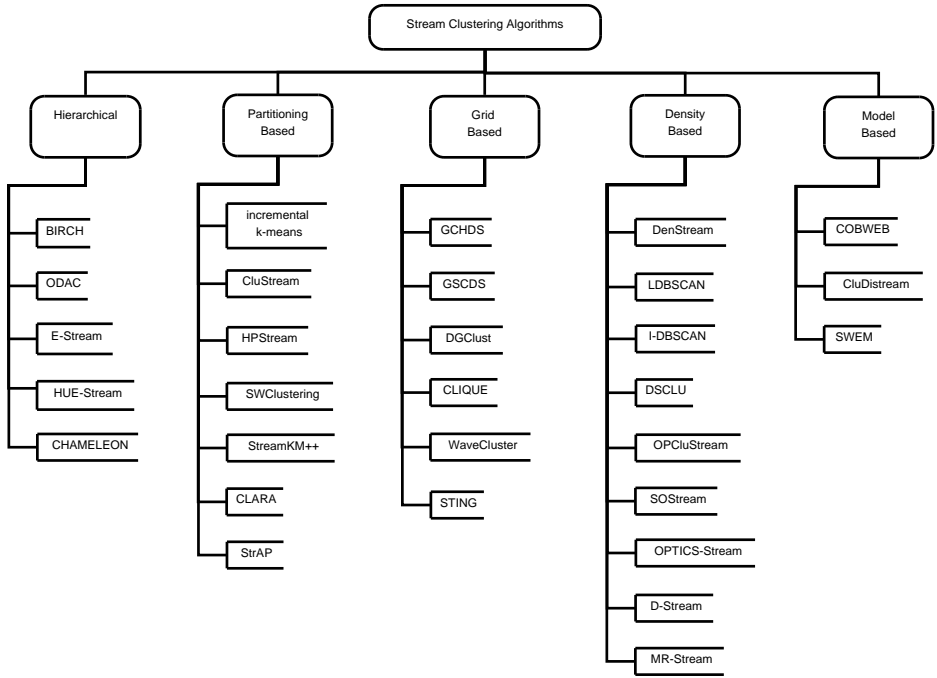}}
	\caption{Classification of data stream clustering algorithms.}
	\label{fig:commonMethods}
\end{figure*}

There exist several data stream clustering algorithms in the literature~\citep{Silva:2013survey, Mousavi:2015:review, ALAM:2016:review, Kumar:2016:review, Ding:2015:review, Ghesmoune:2016:review, RamirezDataPreprocessing, Carnein:2017:review, Keiichi:2016:review, Nguyen:2015:review, Fahy:2018:AntColony, Mahdiraji2009survey, Aggarwal2013chapter}. Data stream clustering algorithms can be categorized following the classification that is used for traditional (batch) clustering algorithms. 
This categorization is given in Figure~\ref{fig:commonMethods} and it consists of five main classes: hierarchical based, partitioning based, density based, grid based and model based clustering. We first give brief information about these categories and related algorithms and we then examine seven most recent data stream clustering algorithms in more detail.

\begin{itemize}
	\item \textit{Hierarchical algorithms} use the dendrogram data structure. Dendrogram is binary tree based, and it is useful to summarize and visualize the data. Hierarchical algorithms are divided in two: agglomerative and divisive. Agglomerative algorithms start with the assumption every instance is a cluster itself, and merge the instances to create clusters step by step. On the other hand, divisive algorithms start assuming a single cluster contains whole data, and divide the clusters into smaller clusters in each step. Hierarchical algorithms have an informative output, which is the dendrogram. However, they have high complexity and they are sensitive to the outliers. Among hierarchical algorithms are BIRCH~\citep{Zhang:1996:Birch}, CHAMELEON~\citep{Karypis:1999:CHAMELEON}, ODAC~\citep{Rodrigues:2006:ODAC}, E-Stream~\citep{2007:E-Stream} and HUE-Stream~\citep{2011:HUE-Stream} ~\citep{Mousavi:2015:review,Kumar:2016:review}. 
	\item \textit{Partitioning based algorithms} split the data instances into a predefined number of clusters, based on similarity (or distance) to the cluster centroids. Number of clusters should be predefined in these algorithms, and only hyper-spherical clusters can be determined. Partitioning based algorithms have an easy implementation in general. StreamLSearch~\citep{Callagha:2002:LSearch}, incremental $k$-means~\citep{2003:incrementalkmeans}, CluStream~\citep{Aggarwal:2003:Clustream}, HPStream~\citep{2004:hpstream}, SWClustering~\citep{Zhou:2008:swclustering}, StreamKM++~\citep{Ackermann:2012:streamKM++}, strAP~\citep{2014:strAP} and CLARA~\citep{2008:clara} are partitioning based algorithms~\citep{Mousavi:2015:review,Kumar:2016:review,Ghesmoune:2016:review}. 
	\item \textit{Grid based algorithms} use grid data structure. The workspace is divided into a number of cells, in a grid structure, and each instance is assigned to a cell. Then, the grid cells are clustered, according to their density. In grid based algorithms, the run time does not depend on input data count. Therefore, grid based algorithms are fast algorithms. Moreover, they are robust to noise and are able to find arbitrary shaped clusters. However, since their complexity depends on the number of the dimensions of the data, grid based algorithms are more suitable for low dimensional data. Furthermore, they need a predefined grid size. GCHDS~\citep{2005:gchds}, GSCDS~\citep{2006:gscds}, DGClust~\citep{Gama:2011:DGClust}, CLIQUE~\citep{1998:clique}, WaveCluster~\citep{2000:WaveCluster} and STING~\citep{1997:STING} are all grid based algorithms~\citep{Mousavi:2015:review}. D-Stream~\citep{Chen:2007:D-Stream} and MR-Stream~\citep{2009:mr-stream} are classified as grid based by~\citet{Ghesmoune:2016:review}, despite being classified as density based by~\citet{Mousavi:2015:review}. 
	\item \textit{Density based algorithms} keep summary of input data in large number of micro-clusters. Micro-cluster is a set of data instances that are very close to each other. Synopsis of a micro-cluster is kept with a feature vector. Then these micro-clusters are merged and formed final clusters according to density reachability and density connectivity concepts. These terms are defined as follows. If the distance between two micro-clusters is less than or equal to the sum of their radii, then they are directly density reachable. If any adjacent two clusters in a set of micro-clusters are directly density reachable, then the set of micro-clusters is density reachable. All micro-clusters that are density reachable to each other, are density connected~\citep{Yin2017improvedClustering}. Density based algorithms are able to find arbitrary shaped clusters and detect number of clusters. They are robust to noise as well. However, several parameters have to be selected and there are problems in finding multi-density clusters. Incremental-DBSCAN~\citep{1998:incrementalDBSCAN}, LDBSCAN~\citep{2006:ldbscan}, DenStream~\citep{2006:denstream}, rDenStream~\citep{2009:rDenStream}, DSCLU~\citep{Namadchian2012DSCLUAN}, OPCluStream~\citep{2012:opclustream}, SOStream~\citep{2012:sostream}, OPTICS-Stream~\citep{2007:opticsstream}, D-Stream~\citep{Chen:2007:D-Stream} and MR-Stream~\citep{2009:mr-stream} are classified as density based~\citep{Mousavi:2015:review,Ghesmoune:2016:review}. 
	\item \textit{Model based algorithms} find the data distribution model that fit best to the input data. One of the important advantages of model based algorithms is their property of noise robustness. However, their performance strongly depends on the selected model. COBWEB~\citep{1996:cobweb}, CluDistream~\citep{2007:cludistream} and SWEM~\citep{2009:swem} are examples of model based algorithms~\citep{Mousavi:2015:review}. 
\end{itemize}
Advantages and disadvantages of clustering algorithms are summarized in Table~\ref{tbl:clusteringClasses} (\cite{Mousavi:2015:review, Mansalis:2018:review, Ghesmoune:2016:review}).

%table
\begin{table}[htbp]
	\caption{Advantages and disadvantages of clustering algorithms based on traditional categorization.}
	\begin{center}
		%\begin{tabular}{l p{3cm} p{3cm}}
		\begin{tabular}{l p{4cm} p{4cm}}
			%\hline
			\textbf{Algorithm} 	& \textbf{Advantages} 			& \textbf{Disadvantages} \\
			\hline
			Hierarchical	& Informative output \newline (dendrogram)		& High Complexity\newline Outlier sensitivity \\
			Partitioning	& Easy implementation & Predefined number of clusters\newline Only hyper-spherical clusters\\
			Grid-based		& Arbitrary shaped clusters\newline Fast execution time\newline Noise robustness & Predefined grid size\newline Only low dimensional data\\
			Density-based	& Arbitrary shaped clusters\newline Noise robustness & Multidensity cluster difficulties\newline Many predefined parameters\\
			Model-based		& Noise robustness  &	Strong dependency on the model \\
			\hline
		\end{tabular}
		\label{tbl:clusteringClasses}
	\end{center}
\end{table}

%\subsection{Most Recent Algorithms}

%Apart from these methods, there exist a number of novel algorithms that are unmentioned in the previous surveys. 
The aforementioned data stream clustering algorithms have already been reviewed in the previous surveys. On the other hand, the algorithms given below have not been analyzed elsewhere, to the best of our knowledge. We give the main flow of the algorithms, show their evaluation results and present the complexity analysis. During the complexity analysis, we ignore the Euclidean distance calculation complexity, which is $O(d)$, because this is the common practice in the literature. Moreover, this calculation is done in every data stream clustering algorithm, thus ignoring it does not change the comparison. However, any other data dimension related complexity is included in the analysis. Not surprisingly, complexity of partitioning based algorithms is a function of $k$ and complexity of density based algorithms is a function of \textit{micro cluster count}.

\begin{figure*}[t]
	\centerline{\includegraphics[width = \textwidth]{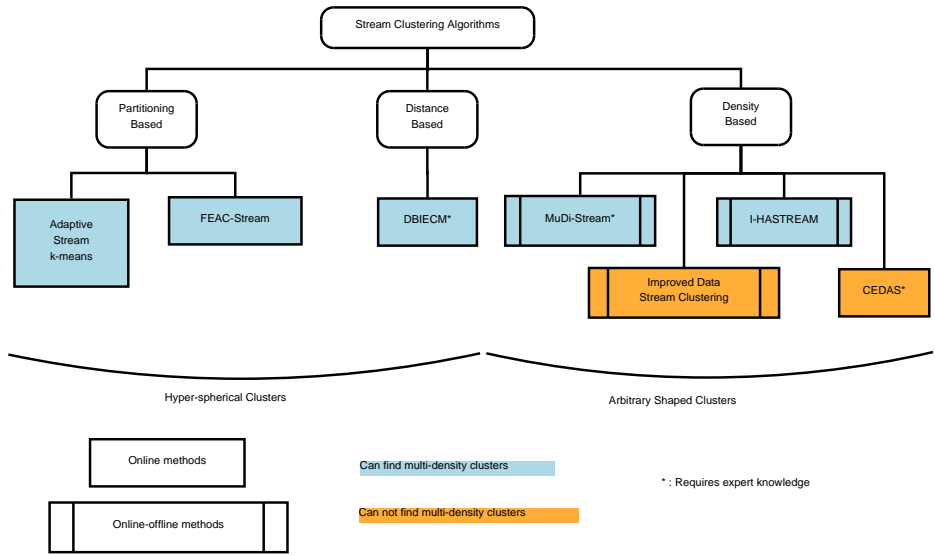}}
	\caption{Recent data stream clustering algorithms.}
	\label{fig:methodsTree}
\end{figure*}

We start with Adaptive Streaming $k$-Means and FEAC-Stream both of which are partitioning based online algorithms. We then examine MuDi-Stream, which is a density based, online-offline algorithm. CEDAS is a density based online algorithm. Improved Data Stream Clustering Algorithm is a density based, online-offline algorithm. DBIECM, the only distance based algorithm is fully online. Note that the previous, classical classification does not include the distance based algorithms, probably, because there are not many examples of distance based algorithms. Finally, we examine I-HASTREAM, a density based hierarchical, online-offline algorithm.
Figure~\ref{fig:methodsTree} shows the main characteristics of the examined algorithms. Although, \textit{ant colony optimization} methods are also being used by a number of data stream clustering algorithms~\citep{Fahy:2018:AntColony}, they are not evaluated at this time. Moreover, being an active research area, there also exist several recent data stream clustering algorithms that are not evaluated in this manuscript~\citep{2020:UDDIN, BEZERRA:2020, KIM:2020}.

\subsection{Adaptive Streaming $k$-Means (2017)}
Adaptive streaming $k$-means is an online, partitioning based data stream clustering algorithm proposed by~\citet{Puschmann:2017:adaptiveStreaming}. In general, partitioning based clustering algorithms need $k$ as an input parameter, and these algorithms have difficulties to adapt concept drift in the input data. In this algorithm,~\citeauthor{Puschmann:2017:adaptiveStreaming} claim to overcome these two main problems.

\begin{algorithm}
	\caption{streamingKMeans (S, $l$)}
	\label{alg:streaminkmeans}
	\begin{algorithmic}[1]
		\Require $S$ : the input data stream
		\Require $l$ : length of data sequence used for initialization
		\State \% Initialization phase
		\For {$candidateCentroids$ \textbf{in} $determineCentroids(l$ number of data instances$)$}
		\State run $kmeans$ with $candidateCentroids$
		\State calculate \textit{silhouette coefficient} of the $kmeans$ result
		\EndFor
		\State keep $centroids$ of the best clustering
		
		\State \% Continuous clustering phase
		\Loop{
			\If {$changeDetected$ on the input stream}
			\State \textit{re-initialize} the algorithm by running again the initialization phase
			\EndIf
			\State run $kmeans$ with last found, best centroids
			\EndLoop}
		
	\end{algorithmic}
\end{algorithm}

Algorithm~\ref{alg:streaminkmeans} shows the main flow of adaptive streaming $k$-means algorithm. $k$-means algorithm and silhouette coefficient calculation function are assumed to be already implemented. The algorithm is composed of two main phases, which are \textit{initialization phase} and \textit{continuous clustering phase}. In the initialization phase, $l$ number of data instances are accumulated. Then groups of candidate centroids are determined at line 2. In function \textit{determineCentroids}, in order to find $k$ and determine candidate centroids, probability density function (PDF) of the data is calculated using kernel density estimation (KDE)~\citep{parzen:1962:kde, rosenblatt:1956:kde}. All directional changes in the shape of PDF curve, are accepted as signs of beginning of a new region. Here the region can be defined as the area between two consecutive directional changes of the PDF curve. Number of regions is considered as a candidate $k$ and centers of these regions are considered as candidate initial centroids. This process is pursued for each feature of the data separately. Because different features generally show different distributions, more than one $k$ values, and different candidate centroids are found. 

After finding candidate $k$ values, clustering is performed for a set of $k$ values where $k \in [k_{min}, k_{min} + k_{max}]$. The \textit{for loop} at lines 2-5 is executed for these values of $k$ and candidate centroids. Clustering results of different $k$ values are compared according to silhouette coefficient, and best $k$ is selected with its corresponding centroids. 

The $loop$ at lines 8-13 runs for continuous clustering phase. Checking for a concept drift (see Section~\ref{subsub:conceptDrift}) is performed at line 9. If no concept drift occurs, clustering of the input data proceeds, at line 12. However if a concept drift exists, $k$ and centroids are recalculated (the algorithm is re-initialized) at line 10, and then clustering continues at line 12 with new $k$ and centroids. 

For concept drift detection, standard deviation and mean of the input data are stored during the execution. The algorithm tracks how these two values change over time and predicts a concept drift according to the change. When a concept drift is predicted, current cluster centroids are no longer valid. In such a case the concept drift is realized at line 9 and a reinitialization is triggered at line 10. Using this mechanism, the algorithm captures the concept drift and adapts itself to the input stream. 

A limitation of this algorithm, being $k$-means based, only hyper-spherical clusters can be detected. Indeed, the authors indicate that $k$-means is used as the underlying clustering technique to clarify the approach, and the concepts of the approach can be applied to different clustering techniques.

\underline{Evaluation:} Adaptive streaming $k$-means algorithm is evaluated against CluStream and DenStream algorithms, according to silhouette coefficient. Artificial datasets with three to five dimensions, that include concept drift, are used as input data streams. Clustering quality improvement of the adaptive streaming $k$-means algorithm is 13\% to 40\% with respect to CluStream. DenStream gives a better clustering quality for one of the datasets, for short time intervals during the execution. However, for the other datasets, clustering quality improvement of the adaptive streaming $k$-means algorithm is up to 280\% with respect to DenStream. Furthermore, the algorithm is evaluated with real traffic data, against the non-adaptive technique, in which, the centroids are never recalculated. Adaptive streaming $k$-means algorithm achieves an improvement up to 31\% in clustering quality when they are compared over the course of one day. When they are compared over the course of one week, clustering quality improvement of the adaptive streaming $k$-means is 12\% on average.

\underline{Complexity Analysis:} Let $l$ be the length of the initial data sequence, and $d$ be the data dimension. Complexity of estimating $k$ for a single dimension is $O(l)$, because this part goes along the PDF and it has a length equal to the data length. Since this estimation is performed for all dimensions, total $k$ estimation complexity becomes $O(d \cdot l)$. After determining initial centroids running $k$-means takes $O(d \cdot k \cdot cs)$ since no iterations of the algorithm are needed, where $cs$ is the number of different centroid sets. Assigning a newly received data instance to the nearest cluster during the online phase is $O(k)$. As a result, total worst case complexity of the algorithm is $O(k)$ + $O(d \cdot l) + O(d \cdot k \cdot cs)$, which equals to $O(d \cdot l)+ O(d \cdot k \cdot cs)$

\subsection{FEAC-Stream (2017)}
Fast evolutionary algorithm for clustering data streams (FEAC-Stream) is an evolutionary algorithm for clustering data streams with a variable number of clusters, proposed by~\citet{Silva:2017:FEAC-Stream}. FEAC-Stream is a $k$-means based algorithm, which estimates $k$ automatically using an evolutionary algorithm. Being fully online, FEAC-Stream does not store synopsis of the data, instead maintains the final clustering result. During the execution, clustering quality is tracked using the Page-Hinkley (PH)~\citep{Page-Hinckley} test and if the quality falls down, the algorithm adjusts itself.

\begin{algorithm}
	\caption{FEAC-Stream (S, $l$, $\lambda$, $\alpha$, $iter$)}
	\label{alg:FEAC-Stream}
	\begin{algorithmic}[1]
		\Require $S$ : the input data stream
		\Require $l$ : length of data sequence used for initialization
		\Require $\lambda$ : decay rate
		\Require $\alpha$ : weight threshold
		\Require $iter$ : $k$-means iteration count
		\State \% Initialization phase
		\State Estimate $k$ with $l$ number of data instances, using \textit{evolutionary algorithm}
		\State state = \textit{normal}
		
		\State \% Continuous clustering phase
		\Loop{
			\State Read next data instance $x$ from data stream $S$
			\State Add $x$ to the nearest cluster
			\State Calculate weight of all clusters
			\State Delete \textit{low weighted} clusters
			\State $PH_{val}$ = Calculate $PH$ test.
			\If {$PH_{val}$ $>$ \textit{warning threshold}}
			\State state = \textit{warning}
			\EndIf
			\If {state \textbf{is} \textit{warning}}
			\State Add $x$ to buffer $B$
			\EndIf
			\If {$PH_{val}$ $>$ \textit{alarm threshold}}
			\State Estimate $k$ with data instances in buffer $B$, using \textit{evolutionary algorithm}
			\State state = \textit{normal}
			\EndIf
			\EndLoop}
		
	\end{algorithmic}
\end{algorithm}

Algorithm~\ref{alg:FEAC-Stream} shows the main flow of FEAC-Stream algorithm. PH test function and the evolutionary algorithm are assumed to be already implemented. The algorithm is composed of two main phases, which are \textit{initialization phase} and \textit{continuous clustering phase}. In the initialization phase, $l$ number of data instances are accumulated. Then $k$ and initial clustering is calculated using an evolutionary algorithm, at line 2 and state is set to normal, at line 3. In this evolutionary algorithm, clustering is performed using $k$-means with a maximum of $iter$ iterations. Simplified silhouette coefficient is used as the fitness function, $k$ is selected randomly such that $k \in [2, \sqrt{l}]$ and initial centroids are also selected randomly from the input data instances.

After clustering the initial $l$ data instances in the initialization phase, the loop at lines 5-21 is executed for continuous clustering phase. When a new data instance is received, it is added to the nearest cluster at line 7. Weight of all clusters are calculated and low weighted clusters are deleted at line 8 and line 9, respectively. After that, PH test is calculated at line 10 and it is compared to warning and alarm threshold values. When PH test value exceeds the warning threshold, the algorithm enters to warning state. In warning state, clustering process continues and received data instances are stored in a buffer, at line 15. If PH test value exceeds alarm threshold, this means a concept drift (see Section~\ref{subsub:conceptDrift}) occurs and current clusters are not valid anymore. When PH test signals an alarm state, it also automatically selects samples from the input data instances that reflects a new partitioning. In such a case, FEAC-Stream clusters the data instances stored in the buffer with the evolutionary algorithm, at line 18 and sets the state back to normal, at line 19. In the evolutionary algorithm, clustering is performed using $k$-means with a maximum of $iter$ iterations. Simplified silhouette coefficient is used as the fitness function, $k$ and initial centroids are specified by the PH test. FEAC-Stream uses damped window model, which is described in Section~\ref{subsubsec:dampedwindow}.

Being $k$-means based, only hyper-spherical clusters can be detected by FEAC-Stream. Moreover, clustering quality of FEAC-Stream strongly depends on the user defined parameters. FEAC-Stream requires three parameters which are length of data sequence used for initialization ($l$), decay rate ($\lambda$) for damped window model and minimum weight threshold ($\alpha$). These parameters strongly affect the clustering quality and they are directly dependent to the input data. Because of that, FEAC-Stream requires an expert knowledge about the input data. Iteration count of $k$-means ($iter$) and generation count of evolutionary algorithm are used as hard coded. Moreover, warning and alarm threshold values of PH test are calculated automatically by the PH test.

\underline{Evaluation:} FEAC-Stream is evaluated against CluStream-OMR$k$, CluStream-B$k$M, StreamKM++-OMR$k$ and StreamKM++-B$k$M, where CluStream and StreamKM++ are the stream clustering algorithms with fixed $k$, while B$k$M and OMR$k$ are $k$ estimating algorithms. Both real and artificial datasets are used for the evaluation. Real datasets are network intrusion detection dataset, forest cover type dataset and localization data for person activity dataset. Adjusted Rand Index (ARI) is used as clustering quality metric in artificial datasets. While all mean ARI results are very close to each other (0.97 - 0.99), FEAC-Stream has the lowest execution time. Its execution time is less by; 25\% than StreamKM++-B$k$M, 58\% than StreamKM++-OMR$k$, 91\% than CluStream-B$k$M and nearly 93\% than CluStream++-OMR$k$. Furthermore, FEAC-Stream successfully reacts to concept drifts and accordingly estimates $k$. For network intrusion detection dataset, simplified silhouette (SS) coefficient is used to compare the clustering quality. Again all algorithms give very good and very close (0.90 - 0.92) SS values and still FEAC-Stream gives the best execution time. Its execution time is less by; 65\% than StreamKM++-B$k$M, 87\% than StreamKM++-OMR$k$, 97\% than CluStream-B$k$M and 98\% than CluStream++-OMR$k$. For the other real datasets as well, algorithms have the same running time ordering. These results also show that, StreamKM++ is faster than CluStream and B$k$M is faster than OMR$k$.

\underline{Complexity Analysis:} Let $l$ be the length of the initial data sequence, $gen$ is generation count of evolutionary algorithm and $iter$ is the iteration count of $k$-means. In the initialization phase, $k$ is randomly selected as $k \in [2, \sqrt{l}]$. Thus, complexity of initialization phase is $O(gen \cdot iter \cdot \sqrt{l})$. Online maintenance of the algorithm requires a complexity of $O(k)$. When a concept drift occurs, the algorithm is reinitialized by running evolutionary algorithm again. However $k$ and centroids are decided by PH test. Therefore, reinitialization requires a complexity of $O(gen \cdot iter \cdot k)$. As a result, total worst case time complexity of FEAC-Stream is $O(k)$ + $O(gen \cdot iter \cdot k)$, which equals to $O(gen \cdot iter \cdot k)$.

\subsection{MuDi-Stream (2016)} 
Multi density data stream clustering algorithm (MuDi-Stream) is a two phase data stream clustering algorithm proposed by~\citet{Amini:2016:MudiStream}. Main objective of MuDi-Stream is to improve the clustering quality on data streams with multi density clusters. Note that density based algorithms usually have problems with clusters of different densities because of the static density threshold they use. MuDi-Stream customizes the density threshold for each cluster and overcomes the problem of multi density clusters. MuDi-Stream is a hybrid algorithm based on both density based and grid based approaches. Input data instances are clustered in a density based approach and outliers are detected using grids. For data synopsis core mini-clusters are used. Core mini-clusters are specialized feature vectors (see Section~\ref{subsec:dataStruct}), they keep weight, center, radius and the maximum distance from an instance to the mean. In the online phase core mini-clusters are created and kept up to date for each new data instance. In the offline phase final clustering is executed over the core mini-clusters.

\begin{algorithm}
	\caption{MuDi-Stream online phase (S, $\alpha$, $\lambda$, $gridGranularity$, $G$)}
	\label{alg:mudiStreamOnline}
	\begin{algorithmic}[1]
		\Require $S$ : the input data stream
		\Require $\alpha$ : density threshold
		\Require $\lambda$ : decay rate
		\Require $gridGranularity$
		\Require $G$ : total density grids for all dimensions
		\State Initialize the grid structure
		\Loop{
			\State Read next data instance $x$ from data stream $S$
			\State $cmc_s$ = Find the nearest $cmc$ to $x$
			\If {$cmc_s$ \textbf{\textit{involve}} $x$}
				\State Add $x$ to $cmc_s$
			\Else
				\State Map $x$ to the gird
				\If {Updated grid is \textit{dense enough}}
					\State Create a $cmc$ from updated grid
				\EndIf
			\EndIf
			\If {It is pruning period}
				%\State Calculate weight of grids
				\State Remove \textit{low weighted} grids
				%\State Calculate weight of $cmc$s
				\State Remove \textit{low weighted} $cmc$s
			\EndIf
		\EndLoop}
		
	\end{algorithmic}
\end{algorithm}

Algorithm~\ref{alg:mudiStreamOnline} shows the main flow of online phase of MuDi-Stream. When a new data instance is received, it is tried to be added to an existing core mini-cluster. For this purpose, the nearest core mini-cluster is found at line 4 and it is checked whether nearest core mini-cluster can involve this data instance or not, at line 5. If the nearest core mini-cluster is \textit{large enough}, the data instance is added to the nearest core mini-cluster at line 6. Otherwise, the data instance is mapped into the gird in the outlier buffer, at line 8. When a data instance is mapped to a grid, density of this grid is checked and if it is \textit{dense enough} (more than the density threshold), a new core mini-cluster is created from this grid i.e. the grid is converted to a core mini-cluster, at line 10. MuDi-Stream prunes both the grids in the outlier buffer and the core mini-clusters periodically. It is checked at line 13 whether it is pruning time or not. If it is pruning time, weight of grids and core mini-clusters are calculated according to current time, and then low weighted grids and core mini-clusters are pruned at line 14 and line 15 respectively. This pruning mechanism is an implementation of damped window model, which is described in Section~\ref{subsubsec:dampedwindow}.

\begin{algorithm}
	\caption{MuDi-Stream offline phase (core mini-clusters)}
	\label{alg:mudiStreamOffline}
	\begin{algorithmic}[1]
		\Require core mini-clusters
		\State Mark all $cmc$s as unvisited
		\Repeat
			\State Randomly choose an unvisited $cmc$, called $cmc_p$
			\State Mark $cmc_p$ as visited
			\If {$cmc_p$ has \textit{neighbors}}
				\State Create new final cluster $C$
				\State Add $cmc_p$ to $C$
				\State Add \textit{neighbors} of $cmc_p$ to $C$
				\For {\textbf{each} $cmc$ in $C$}
					\If {$cmc$ is unvisited}
						\State Mark $cmc$ as visited
						\State Add \textit{neighbors} of $cmc$ to $C$
					\EndIf
				\EndFor
			\Else
				\State Mark $cmc_p$ as \textit{noise}
			\EndIf
			
		\Until All $cmc$s are visited
	\end{algorithmic}
\end{algorithm}

Algorithm~\ref{alg:mudiStreamOffline} shows the main flow of offline phase of MuDi-Stream. Initially all core mini-clusters are marked as unvisited, at line 1. After that, inside a loop, an unvisited core mini-cluster is randomly chosen at line 3 and marked as visited at line 4. If this core mini-cluster has no neighbors, it is marked as noise at line 16. If it has neighbors, a new final cluster is created with this core mini-cluster and its neighbors, at lines 6-8. After that, each unvisited core mini-cluster in the new created final cluster is marked as visited and its neighbors are added to the same final cluster, at lines 9-14. This loop continues until all core mini-clusters are marked as visited. 

Damped window model is used, and arbitrary shaped, multi density clusters can be detected by MuDi-Stream. Moreover, MuDi-Stream is able to handle concept drift (see Section~\ref{subsub:conceptDrift}), noise and outliers. However it is not suitable for high dimensional data, which makes the processing time longer, because of the grid structure. Furthermore, clustering quality of MuDi-Stream strongly depends on input parameters density threshold ($\alpha$), decay rate ($\lambda$) for damped window model and \textit{grid granularity}. These parameters require an expert knowledge about the data.

\underline{Evaluation:} MuDi-Stream is tested with two real (network intrusion detection and Landsat satellite) and six artificial datasets. It is compared to DenStream on a data stream with concept drifts, a multi density dataset and a multi density data stream with concept drifts. MuDi-Stream outperforms DenStream on all three types of input data, according to clustering quality (Purity, Normalized Mutual Information (NMI), Rand Index (RI), Adjusted Rand Index (ARI), Folkes and Mallow index (FM), Jaccard Index and F-Measure). Clustering quality improvement of MuDi-Stream is 10\% to 100\% with respect to DenStream, on different datasets.

\underline{Complexity Analysis:} MuDi-Stream performs a linear search on core mini-clusters for each new data instance. Complexity of this linear search is $O(c)$ where $c$ is the number of core mini-clusters. If the new data instance cannot be merged into existing core mini-clusters, it is mapped to the grid. Let $G$ be total density grids for all dimensions, which is exponential to the number of dimensions. Space complexity of the grid is $O(log\ G)$ because the scattered grid are pruned during the execution. Moreover, time complexity of mapping a data instance to the grid is $O(log\ log\ G)$ because the list of the grids is maintained as a tree. During the pruning, all core mini-clusters and grids are examined. This makes time complexity of pruning $O(c)$ for core mini-clusters and $O(log\ G)$ for grids. As a result, the overall time complexity of MuDi-Stream is $O(c)$ + $O(log\ log\ G)$ + $O(c)$ + $O(log\ G)$, which equals to $O(c)$ + $O(log\ G)$.

\subsection{CEDAS (2016)}
Clustering of evolving data streams into arbitrarily shaped clusters (CEDAS) is a fully online data stream clustering algorithm proposed by~\citet{Hyde:2017:Cedas} CEDAS is a density based algorithm designed for clustering data streams with concept drifts (see Section~\ref{subsub:conceptDrift}), into arbitrary shaped clusters. Damped window model (see Section~\ref{subsubsec:dampedwindow}) is employed with a linear decay function instead of an exponential one. CEDAS keeps synopsis of the data in micro-clusters and creates a graph structure with the micro-clusters that surpass a user defined threshold. Graph structure, where nodes are the micro-clusters and edges are the connectivity between micro-clusters, keeps the up to date final clustering results.

\begin{algorithm}
	\caption{CEDAS (S, $\alpha$, $\lambda$, $r_0$)}
	\label{alg:cedas}
	\begin{algorithmic}[1]
		\Require $S$ : the input data stream
		\Require $\alpha$ : density threshold
		\Require $\lambda$ : decay rate
		\Require $r_0$ : micro-cluster radius
		\State Initialize the micro-cluster structure
		\Loop{
			\State Read next data instance $x$ from data stream $S$
			\State $dis_{min}$ = Find the distance from $x$ to the nearest micro-cluster center
			\If {$dis_{min}$ $<$ $r_0$}
				\State Add $x$ to the nearest micro-cluster
				\State Energy of the updated micro-cluster = 1
			\Else
				\State Create new micro-cluster with $x$
				\State Energy of the new micro-cluster = 1
			\EndIf
			\State Reduce energy of all micro-clusters by $\lambda$
			\State Remove \textit{negative energy} micro-clusters
			\If {micro-clusters are \textit{changed}}
				\State Update graph structure with micro-clusters that surpass $\alpha$
			\EndIf
		\EndLoop}
		
	\end{algorithmic}
\end{algorithm}

Algorithm~\ref{alg:cedas} shows the main flow of CEDAS. When a new data instance is received, it is tried to be added to an existing micro-cluster. For that purpose, the distance from new data instance to the nearest micro-cluster is found at line 4 and it is checked whether this distance is less than the micro-cluster radius ($r_0$) or not, at line 5. Micro-cluster radius is a user defined, static parameter. If the distance is less than the radius, the data instance is added to the nearest micro-cluster, at line 6, and energy of this micro-cluster is set to $1$ at line 7. Otherwise, a new micro cluster is created with this data instance, at line 9, and energy of the new micro-cluster is set to 1, at line 10. Energy of micro-clusters linearly fades on every cycle, with an amount of decay rate ($\lambda$), at line 12. The micro-clusters whose energy drop below zero are removed at line 13. Lastly, the graph structure is updated with the micro-clusters that surpass the density threshold ($\alpha$), at line 15. Removed micro-clusters are removed from the graph structure also, and micro-clusters reached the density threshold ($\alpha$) added to the graph structure. Therefore, CEDAS creates final clustering results as fully online.

CEDAS is suitable for high dimensional data under favor of maintaining a graph structure where nodes are the micro-clusters and edges are the connectivity between micro-clusters. However, clustering quality of CEDAS strongly depends on the user defined parameters. CEDAS requires three parameters which are decay rate ($\lambda$), micro-cluster radius ($r_0$) and minimum density threshold ($\alpha$). These parameters strongly affect the clustering quality and they are directly dependent to the input data. Because of that, CEDAS requires an expert knowledge about the input data.

\underline{Evaluation:} CEDAS is tested with a data stream consisting of two Mackey-Glass time series, to see how it deals with concept drift, cluster separation, cluster merging and noise over time. Moreover, it is compared to CluStream and DenStream according to complexity, processing speed, cluster quality and memory efficiency. CEDAS, CluStream and DenStream are also compared with high dimensional data according to speed and accuracy. CEDAS successfully deals with concept drift. Noise negatively affects the clustering quality, however results are claimed to be still acceptable. Time measurements show that CEDAS is quite suitable for high dimensional data. Firstly CEDAS is compared against only online phases of DenStream and CluStream. For data with less than 10 dimensions, CEDAS is the slowest one. However, processing time of CEDAS stays nearly constant up to 10,000 dimensions. CluStream becomes slower than CEDAS after 10 dimensions and it consumes nearly 300 times more than CEDAS for 6,000 dimensions. DenStream is faster than CEDAS up to 200 dimensions. For more than 200 dimension, DenStream becomes slower than CEDAS and consumes nearly 2 times more than CEDAS for 6,000 dimensions. After that, CluStream and DenStream are run with a frequent offline phase, to generate near real time final clustering. In this situation CEDAS is the fastest algorithm for both low and high dimensional data. For 5 dimensional data, DenStream consumes 40 times and CluStream consumes 75 times more than CEDAS. For very high dimensional data, time consumption of DenStream grows faster than the others. When the data dimension is 3,000 CluStream consumes nearly 100 times and DenStream consumes nearly 650 times more than CEDAS. The other main advantage of CEDAS is memory efficiency. During the execution, DenStream reaches up to 800 micro-clusters at certain times, while CEDAS reaches up to 100 micro-clusters.

\underline{Complexity Analysis:} For each new data instance, CEDAS performs a linear search on the micro-clusters. Complexity of this linear search is $O(c)$ where $c$ is the number of micro-clusters. After that, energy of each micro-cluster is reduced, which also requires an $O(c)$ complexity. The last step, which updates the graph structure, is executed only when a new micro-cluster is created or removed. In worst case, all micro-clusters are visited, so worst case time complexity of this step is again $O(c)$. Therefore, the overall time complexity of CEDAS is $O(c)$.

\subsection{Improved Data Stream Clustering Algorithm (2017)}
Improved data stream clustering algorithm is a two phase, density based algorithm that is suitable for arbitrary shaped clusters, proposed by~\citet{Yin2017improvedClustering}. Main characteristic of this algorithm is adjusting threshold values automatically, according to the input data. This feature gets rid of the requirement of expert knowledge about the input data.

\begin{algorithm}
	\caption{Improved data stream clustering online phase (S, $l$, $\lambda$)}
	\label{alg:idsClusteringOnlineShort}
	\begin{algorithmic}[1]
		\Require $S$ : the input data stream
		\Require $l$ : length of data sequence used for initialization
		\Require $\lambda$ : decay rate
		\State \% Initialization phase
		\State Run DBSCAN on $l$ number of data instances
		
		\State \% Continuous clustering phase
		\Loop{
			\State Read next data instance $x$ from data stream $S$
			\State Add $x$ to the nearest \textit{major micro-cluster} \textbf{OR}
			\State Add $x$ to the nearest \textit{critical micro-cluster} \textbf{OR}
			\State Create a new \textit{micro-cluster} with $x$
			\If {It is pruning period}
				\State Remove \textit{low weighted} major micro-clusters
				\State Remove \textit{low weighted} critical micro-clusters
			\EndIf
			\EndLoop}
		
	\end{algorithmic}
\end{algorithm}

Algorithm~\ref{alg:idsClusteringOnlineShort} shows the main flow of online phase of improved data stream clustering algorithm. DBSCAN algorithm is assumed to be already implemented. The algorithm is composed of two main phases, which are \textit{initialization phase} and \textit{continuous clustering phase}. In the initialization phase, $l$ number of data instances are accumulated and clustered using DBSCAN, at line 2. Major micro-clusters and critical micro-clusters are created as output of DBSCAN algorithm. Major micro-clusters have high densities and will be included in the final clustering process. Critical micro-clusters have low densities and treated as potential outliers. In the continuous clustering phase, when a new data instance is received, it is tried to be added to the nearest major micro-cluster, at line 6. If nearest major micro-cluster is not suitable, this time the new data instance is tried to be added to the nearest critical micro-cluster, at line 7. If neither of them is suitable, a new micro-cluster is created with the new data instance, at line 8. Damped window model (see Section~\ref{subsubsec:dampedwindow}) is used and low weighted major and critical micro-clusters are removed periodically, at line 10 and line 11 respectively. Threshold values of major and critical micro-clusters are global parameters in the algorithm, instead of being specific to each micro-cluster. However they are dynamic parameters and continuously updated during the execution.

\begin{algorithm}
	\caption{Improved data stream clustering offline phase (micro-clusters)}
	\label{alg:idsClusteringOffline}
	\begin{algorithmic}[1]
		\Require micro-clusters
		\State Mark all $mc$s as unvisited
		\Repeat
		\State Randomly choose an unvisited $mc$, called $mc_p$
		\If {$mc_p$ \textbf{is} \textit{major micro-cluster}}
		\State Find all micro-clusters \textit{density reachable} to $mc_p$
		\State Create a final cluster by them.
		\ElsIf {$mc_p$ \textbf{is} \textit{critical micro-cluster}}
		\State Continue the next cycle
		\EndIf
		\Until All $mc$s are visited
		
	\end{algorithmic}
\end{algorithm}

Algorithm~\ref{alg:idsClusteringOffline} shows the main flow of offline phase of improved data stream clustering algorithm. Initially all micro-clusters are marked as unvisited, at line 1. After that, inside a loop, an unvisited micro-cluster is chosen randomly at line 3. If the selected micro-cluster is a major micro-cluster, all micro-clusters that are density reachable to this micro-cluster are found and a new final cluster is created by them, at line 5 and line 6. If the selected micro-cluster is a critical micro-cluster, then the execution continues with the next cycle, at line 8. When all micro-clusters are visited, the offline phase completes. The term density reachable is defined as follows. If the distance between a micro-cluster and another major micro-cluster is less than or equal to the sum of their radii, then they are directly density reachable. If any adjacent two clusters in a set of micro-clusters are directly density reachable, then the set of micro-clusters is density reachable~\citep{Yin2017improvedClustering}.

\underline{Evaluation:} Improved data stream clustering algorithm is evaluated against DenStream algorithm, using the network intrusion detection dataset. Clustering quality improvement of the improved data stream clustering algorithm is 2\% to 7\% with respect to DenStream. Moreover,~\citet{Yin2017improvedClustering} claims that this algorithm has a better time and spatial complexity, compared with traditional clustering algorithms, however no measurement results are shared. 

\underline{Complexity Analysis:} Let $l$ be the length of the initial data sequence. Complexity of the initialization equals to complexity of DBSCAN, which is $O(l\cdot log \ l)$ in average and $O(l^2)$ in worst case. In the continuous clustering phase, a linear search is performed on micro-clusters for each new data instance. Complexity of this linear search is $O(c)$ where $c$ is the number of micro-clusters. When it is pruning period, pruning task is executed for each micro-cluster one by one and this also requires a complexity of $O(c)$. Therefore, the total worst case complexity is $O(c)$ + $O(c)$, which equals to $O(c)$.

\subsection{DBIECM (2017)}
DBIECM is an online, distance based, evolving data stream clustering algorithm proposed by~\citet{Zhang:2017:dbiecm}. DBIECM is the only example of distance based clustering algorithms in this survey. DBIECM is an improved version of Evolving Clustering Method (ECM)~\citep{Qun:2001:Ecm}. Davies Bouldin Index (DBI) is used as the evaluation criteria, instead of shortest distance.

\begin{algorithm}
	\caption{DBIECM (S, $r_0$)}
	\label{alg:dbiec}
	\begin{algorithmic}[1]
		\Require $S$ : the input data stream
		\Require $r_0$ : max cluster radius
		\State Initialize the cluster structure
		\Loop{
			\State Read next data instance $x$ from data stream $S$
			\State $dis_i$ = Find the distance from $x$ to all cluster centers $C_i$, $i \in [1, k]$
			\If {$dis_i$ $<$ radius of $C_i$}
				\State Add $x$ to $C_i$
			\ElsIf {$dis_i$ $>$ $r_0$ for all $i \in [1, k]$}
				\State Create new micro-cluster with $x$
			\Else \  \% There exist clusters such that radius of $C_i$ $<$ $dis_i$ $<$ $r_0$
				\State Find all clusters such that radius of $C_i$ $<$ $dis_i$
				\State Add $x$ to the best cluster, according to DBI
			\EndIf
		\EndLoop}
		
	\end{algorithmic}
\end{algorithm}

Algorithm~\ref{alg:dbiec} shows the main flow of DBIECM. When a new data instance $x$ is received, an attempt is made to add the new data instance to an existing cluster. For this purpose, the distances between $x$ and all clusters are calculated. If radius of any cluster is greater than or equal to its distance to $x$, then $x$ is added to this cluster, as indicated at line 6. If the distance from $x$ to any cluster is greater than maximum cluster radius $r_0$, which is a user defined, static parameter, then a new cluster is created with $x$, at line 8. Otherwise, if there exist any clusters such that their radii are less than their distance to $x$, then $x$ is added to all of these clusters one by one and DBI of the results are calculated separately. $x$ is added to the cluster that gives the least DBI, which means the best clustering.

DBIECM requires the maximum cluster radius as a parameter. This parameter directly affects the final cluster count and consequently the clustering quality. Maximum cluster radius strongly depends on the input data and requires an expert knowledge about the data. Being distance based, DBIECM can detect only hyper-spherical clusters. DBIECM does not employ any time window model, thus no input data instance out dates, all input data exist in the final clustering. Moreover, no outlier detection mechanism is implemented. However, it is possible to specify an outlier threshold value and mark the clusters with low cardinality as outliers. 

\underline{Evaluation:} DBIECM is evaluated against ECM, with Iris, Wine, Seeds, Glass and Breast Cancer datasets, from UCI machine learning database. Both of the algorithms are run with the same maximum cluster radius parameter. Firstly, three different radius values are tried, and their direct impact on the resultant cluster number is observed. This shows the importance of the expert knowledge for radius selection. Moreover, clustering quality is compared according to objective function value, DBI, accuracy and purity. For these tests, radius value is selected according to the correct cluster number. DBIECM achieve up to 43\% better DBI, up to 33\% better accuracy and up to 11\% better purity values than ECM.

\underline{Complexity Analysis:} When a new data instance is received, a linear search is performed on clusters. Complexity of this linear search is $O(k)$. Pairwise distances between all clusters are used for DBI calculation, thus DBI calculation requires a complexity proportional to $O(k^2)$. When there exist more than one candidate clusters for the new data instance, the instance is added to all of them one by one and DBI is calculated accordingly. This requires a complexity proportional to $O(k^3)$. Therefore, although the average complexity of DBIECM depends on the input data, the total worst case complexity is $O(k)$ + $O(k^3)$ which equals to $O(k^3)$.

\subsection{I-HASTREAM (2015)}
I-HASTREAM is a two phase, adaptive, density based hierarchical, data stream clustering algorithm proposed by~\citet{Hassani:2015:I-HASTREAM, Hassani:2016:I-HASTREAM}. I-HASTREAM is an improved version of HASTREAM~\citep{Hassani:2014:HASTREAM}. In the online phase, synopsis of the data is created as micro-clusters. In the offline phase, micro-clusters are maintained in a graph structure as a minimum spanning tree and hierarchical clustering is employed for the final clustering. Main contributions of I-HASTREAM are to perform the final clustering on a minimum spanning tree and to incrementally update the minimum spanning tree according to the changes in the micro-clusters, instead of generating it from scratch. Both of these contributions are related to the offline phase. For I-HASTREAM and its ancestor HASTREAM~\citep{Hassani:2014:HASTREAM} no algorithmic details are specified about the online phase, instead, it is stated that any micro-cluster model can be employed. For evaluation purpose, HASTREAM employs online phases of \textit{DenStream} and \textit{ClusTree} algorithms and these results are presented by~\citet{Hassani:2014:HASTREAM}.

\begin{algorithm}
	\caption{I-HASTREAM offline phase (micro-clusters, $\alpha$)}
	\label{alg:ihastreamOffline}
	\begin{algorithmic}[1]
		\Require micro-clusters
		\Require $\alpha$ : weight threshold
		\State $MST$ = Update minimum spanning tree($MST$, micro-clusters)
		\State $HC$ = Employ hierarchical clustering($MST$, $\alpha$)
		\State Extract final clustering($HC$)
	\end{algorithmic}
\end{algorithm}

Algorithm~\ref{alg:ihastreamOffline} shows main flow of offline phase of I-HASTREAM. The minimum spanning tree is updated according to the changes in the micro-clusters at line 1, and a hierarchical clustering on the minimum spanning tree is employed at line 2. As result of hierarchical clustering, a dendrogram is created. Final clustering is performed according to this dendrogram, at line 3. 

\underline{Evaluation:} Four variants of I-HASTREAM (with different parameters) are evaluated against HASTREAM, MR-Stream and DenStream, using network intrusion detection dataset and the physiological dataset. Purity and Cluster Mapping Measure (CMM)~\citep{Kremer:2011:CMM} are used as evaluation criteria. One of the I-HASTREAM variants gives up to 25\% better purity values than DenStream in network intrusion detection dataset. Its result is also up to 10\% better than other versions of I-HASTREAM and HASTREAM. In the physiological dataset, the same variant of I-HASTREAM gives the best CMM and purity values in general. HASTREAM and I-HASTREAM have very close CMM values and both of them outperforms DenStream with up to 30\% better CMM values. For purity, again I-HASTREAM has the best values in general and it outperforms both DenStream and MR-Stream with up to 15\% better purity values. When we look at the execution time comparison of the algorithms, I-HASTREAM is more than five times faster than DenStream.

\underline{Complexity Analysis:} Because no algorithmic details are specified about the online phase, we could not analyze complexity of I-HASTREAM.

\section{Comparison of the Algorithms}
\label{sec:methodComparison}

\begin{table*}[t]
	\centering
	\caption{Comparison of recent data stream clustering algorithms.}
	\begin{tabular}{p{2cm} p{0.7cm} p{1.3cm} p{1cm} p{1.2cm} p{0.8cm} p{1.2cm}}
		\textbf{Algorithm} & \textbf{Year} & \textbf{Base Algorithm} & \textbf{Phases} & \textbf{Window Model} & \textbf{Cluster Count} & \textbf{Cluster Shape} \\
		\hline
		Adaptive Streaming $k$-Means & 2017 & Partitioning based & Online & Sliding & Auto & Hyper-spherical  \\
		\hline
		FEAC-Stream & 2017 & Partitioning based & Online & Damped & Auto & Hyper-spherical\\
		\hline
		MuDi-Stream & 2016 & Density based & Online-offline & Damped & Auto & Arbitrary \\
		\hline
		CEDAS & 2016 & Density based & Online & Damped & Auto & Arbitrary \\
		\hline
		Improved Data Stream Clustering & 2017 & Density based & Online-offline & Damped & Auto & Arbitrary \\
		\hline
		DBIECM & 2017 & Distance based & Online & None & Auto & Hyper-spherical \\
		\hline
		I-HASTREAM & 2015 & Density based & Online-offline & Damped & Auto & Arbitrary \\
		\hline
	\end{tabular}

	\label{tbl:comparison}
\end{table*}

\begin{table*}[t]
	\centering
	\caption{Comparison of recent data stream clustering algorithms (continued from Table~\ref{tbl:comparison}).}
	\begin{tabular}{p{2cm} p{1.25cm} p{2cm} p{1.5cm} p{1.5cm} p{1.5cm}}
		\textbf{Algorithm} & \textbf{Multi Density Clusters} & \textbf{High Dimensional Data} & \textbf{Outlier Detection} & \textbf{Drift Adaption} & \textbf{Expert Knowledge} \\
		\hline
		Adaptive Streaming $k$-Means & Yes & Suitable & No & Yes & No  \\
		\hline
		FEAC-Stream & Yes & Suitable & Yes & Yes & No \\
		\hline
		MuDi-Stream & Yes & Not suitable & Yes & Yes & Required\\
		\hline
		CEDAS & No & Suitable & Yes & Yes & Required \\
		\hline
		Improved Data Stream Clustering & No & Suitable & Yes & Yes & No \\
		\hline
		DBIECM & Yes (not multi size) & Suitable & No & Yes & Required \\
		\hline
		I-HASTREAM & Yes & Suitable & Yes & Yes & No \\
		\hline
	\end{tabular}
	\label{tbl:comparison2}
\end{table*}

As common characteristics of seven data stream clustering algorithms given in Section~\ref{sec:methods}, all of them predict number of clusters themselves and they are all able to adopt concept drift in the data streams. All but MuDi-Stream are suitable for high dimensional data. The reason MuDi-Stream is not suitable for high dimensional data is that, it uses a grid based approach for outlier detection. When the data are high dimensional, the number of empty grids increases and the execution time gets higher. 

Adaptive Streaming $k$-means and FEAC-Stream are both $k$-means based (partitioning based) algorithms. DBIECM is distance based and the others are density based algorithms. Distance based approaches are similar to density based approaches, however they do not have a density threshold, instead they have maximum cluster radius threshold. 

In general, density based algorithms have problem about finding clusters with different densities, because of the static density threshold. However, MuDi-Stream and I-HASTREAM have improvements for this problem and they successfully adopt the density threshold to each cluster separately. This makes them able to find multi-density clusters. Adaptive Streaming $k$-means and FEAC-Stream, being partition based algorithms, are also able to find clusters with different densities. DBIECM is successful for multi density clusters, but not for multi size clusters. It has a static maximum cluster radius threshold and this is a problem for clusters with different sizes. As a result, CEDAS and Improved Data Stream Clustering algorithm are not able to find multi density clusters, but the others are. Furthermore, all density based algorithms are able to find arbitrary shaped clusters, while partitioning and distance based algorithms are limited with hyper-spherical clusters. 

For Adaptive Streaming $k$-means and DBIECM, no outlier detection mechanism is mentioned. However, it is possible to define an outlier threshold and to mark the clusters have less cardinality than the threshold as outliers, for both algorithms. The other algorithms already have outlier detection mechanisms. 

Up to the recent years, most of data stream clustering algorithms were online-offline algorithms. A synopsis of the data is employed in the online phase and the final clusters are generated in the offline phase. In this type of algorithms, offline phase is executed periodically or upon user request. Therefore, final clustering results are obtained with a latency and they are not up to date most of the times. However, there exist several recent fully online algorithms in the literature. Fully online algorithms maintain the final clustering results up to date. Therefore, users get the results with no latency. CEDAS, Adaptive Streaming $k$-means, FEAC-Stream and DBIECM are online algorithms, while MuDi-Stream, Improved Data Stream Clustering and I-HASTREAM are online-offline algorithms. 

Damped window model is the most popular time window model among data stream clustering algorithms. On the other hand, DBIECM does not use any time window model. Moreover, Adaptive Streaming $k$-means uses sliding window model. All other mentioned algorithms use damped window model. 

Finally, clustering quality of MuDi-Stream, CEDAS and DBIECM is strongly sensitive to the input parameter \textit{threshold} value. It directly affects the number of clusters and accordingly the clustering quality. Selecting a proper threshold value requires an expert knowledge about the input data. Therefore, for successful results of MuDi-Stream, CEDAS and DBIECM, it is necessary to have prior information about characteristics of the input data. Table~\ref{tbl:comparison} and Table~\ref{tbl:comparison2} show the comparison summary of examined data stream clustering algorithms and Figure~\ref{fig:methodsTree} shows their main characteristics.

In conclusion, Adaptive Streaming $k$-means, FEAC-Stream and DBIECM have limitations about the cluster shape; they are able to find only hyper-spherical clusters. MuDi-Stream is not suitable for high dimensional data because of its grid based outlier detection mechanism. CEDAS and Improved Data Stream Clustering algorithm can not be used for clusters with different densities and DBIECM can not be used for clusters with different radii. Finally, an expert knowledge about the input data and the clusters is required for MuDi-Stream, CEDAS and DBIECM. I-HASTREAM claims to have no limitations, however no algorithmic details are specified for online phase of it. It is stated that online phases of \textit{DenStream} and \textit{ClusTree} are employed instead.

\section{Open Problems}
\label{sec:openProblems}
There exist several open problems about data stream clustering. Here, we indicate the most notable open problems and describe them briefly.
\begin{itemize}
	\item \textbf{Finding $k$:} Finding $k$ is still an open problem, especially for partitioning based algorithms. There exist some recent methods for this purpose, however none of them is widely accepted and well matured yet. For density based algorithms, determining $k$ is easier, however parameters that depend on domain knowledge are necessary. If cluster characteristics such as density and minimum allowable gap between clusters are known \textit{a priori}, current algorithms are then able to detect $k$; however, in most cases, knowledge about input data is not available before the execution and it may not be possible to specify parameters that are valid for all clusters. For example, multi-density clusters require different density thresholds and multi-size clusters require different distance thresholds. Determining such parameters is another open problem by itself. Moreover, concept drift, which may invalidate data specific parameters, is very common in data streams. Therefore, finding a $k$ estimation method that adopts to changes in both $k$ and cluster characteristics is a challenge. Such a method should react to concept drift fast, adopt the new data distribution with minimum quality loss and estimate $k$.
	\item \textbf{Parameter Requirements:} Current data stream clustering algorithms require parameters such as $k$, density threshold, distance threshold, decay rate and window length. Such parameters are very sensitive to the input data and they directly affect the clustering quality. It is a challenge to automatically specify these parameters without domain knowledge, manage them for each cluster separately, and update them according to the data characteristics.
	\item \textbf{Evaluation Criteria:} There is no \textit{de facto} evaluation criteria for data stream clustering. Traditional evaluation methods are used for stream clustering results. Defining a new evaluation metric that is suitable for data streams might contribute to this field and inspire interest.
	\item \textbf{Benchmark Data:} There is a lack of high quality benchmark data to use in data stream clustering algorithms. One of the most popular datasets for stream clustering is the forest cover type dataset and it is not even a stream data. Artificial and real datasets that include concept drift, outliers and class labels, are necessary for benchmarking purposes in data stream clustering field. Generating and collecting such artificial and real stream datasets and popularizing them is a challenge.
	\item \textbf{Experimental Comparison Environment:} There is not a system that runs more than one data stream clustering algorithms at the same time, feeds them in the same way, and compares their execution performance and clustering quality. 
	\item \textbf{Different Data Types:} Handling different data types is another challenging task in data stream clustering. Most of the stream clustering algorithms work with quantitative features and define the similarity based on euclidean distance. Current data structures that keep the data synopsis are also specialized for quantitative features. There exists a lack of clustering algorithms that work with categorical data. It is common to convert categorical data to quantitative data and use existing algorithms. 
	\item \textbf{Performance Improvements:} Any performance improvements is always welcome, since the number of connected devices is increasing and the data generated by them are scaling up and accelerating every day. This situation requires a continuous performance improvement in data stream clustering algorithms. It is possible to improve the performance by using \textit{parallel programming} and \textit{edge computing}. However in this study, we focus on processing where the whole data is gathered and processed directly on a single processor.
\end{itemize}
Concept drift is a data stream specific and it generates several challenges. The number of clusters, cluster densities, sizes and shapes may change over time due to concept drift. The problems of traditional clustering become continuous problems for stream clustering.

\section{Popular Data Repositories and Datasets}
\label{sec:dataReposSets}
\subsection{Data Repositories}
There exist several stream data resources on the internet. Moreover, it is common to use traditional datasets as streams or to generate artificial data streams. Traditional datasets are generally read by order and treated as streams for testing and benchmarking purposes. We mention the stream data sources in this section. Data streams in Stream Data Mining Repository (see Section~\ref{subsubsec:StreamDataMinngRepo}) and MOA (see Section~\ref{subsubsec:moa}) already have true class labels. However, Citi Bike System Data (see Section~\ref{subsubsec:citibike}) does not possess explicitly a class label. One should decide how to employ the data and then assign accordingly the class labels. Moreover, National Weather Service Public Alerts (see Section~\ref{subsubsec:nationalweather}) and Meetup RSVP Stream (see Section~\ref{subsubsec:meetup}) have several features that can be used as class labels.

\subsubsection{Citi Bike System Data}
\label{subsubsec:citibike}
\citet{CitiBike} is a public bicycle sharing system. It is composed of 750 stations and 12,000 bikes. Citi Bike publicly publishes real time system data in~\citeauthor{CitiBike:SystemData} which includes system information, station information, free bike status etc. in a json structure. Moreover, Citi Bike also publishes trip histories, daily ridership and membership data, and monthly operating reports stored as data streams.

\subsubsection{Meetup RSVP Stream}
\label{subsubsec:meetup}
\citet{Meetup} is a website providing membership software, allowing its users to schedule events using a common platform. Meetup has an invitation response mechanism in which the invitees click to RSVP button and enter their responses. Meetup publicly publishes these RSVP responses as a stream~\citep{Meetup:RSVP}, which is suitable for data stream clustering.

\subsubsection{National Weather Service Public Alerts}
\label{subsubsec:nationalweather}
\citet{NWS} creates public alerts, watches, warnings, advisories, and other similar products in the Common Alerting Protocol (CAP) and Atom Syndication Format (ATOM)~\citep{NWS:alerts}. These are data streams and they can be used for data stream clustering studies.
	
\subsubsection{Stream Data Mining Repository}
\label{subsubsec:StreamDataMinngRepo}
Stream Data Mining Repository is a public repository by~\citet{StreamDataRepo} holding four different stream datasets, which are Sensor Stream (2,219,803 instances, 5 features, and 54 clusters), Power Supply Stream (29,928 instances, 2 features, and 24 clusters), Network Intrusion Detection 10\% Subset (494,021 instances, 41 features, and 23 clusters) and Hyper Plane Stream (100,000 instances, 10 features, and 5 clusters).

\subsubsection{MOA}
\label{subsubsec:moa}
\citeauthor{Moa}~\citep{Bifet:MOA} is a popular open source framework for data stream mining. MOA includes 4 different datasets which are suitable for data stream processing. Moreover, it also includes a number of classes to generate artificial data streams. There exist several studies in the literature that use MOA as a data source. More information about MOA is available in Section~\ref{subsec:moa} and artificial data stream generation classes of MOA are listed in Section~\ref{subsubsec:artificalStreams}.

\subsubsection{Other Repositories}
Some other data repositories are listed here.
\begin{itemize}
 \item Real World Data in Real Time API : https://www.hooksdata.io/
 \item New York City Open Data : https://opendata.cityofnewyork.us/
 \item Registry of Open Data on AWS : https://registry.opendata.aws/
 \item Twitter Data :\\ https://developer.twitter.com/en/docs/tutorials/consuming-streaming-data
 \item AirNow Air Quality Observations : https://docs.airnowapi.org/
 \item National Wind Technology Center (NWTC) :\\ https://data.nrel.gov/submissions/33
 \item Solar Radiation Research Laboratory (SRRL) :\\ https://data.nrel.gov/submissions/7
 \item Awesome Public Datasets :\\ https://github.com/awesomedata/awesome-public-datasets
\end{itemize}

\subsection{Popular Datasets}
It is very common to use artificial datasets in data stream clustering for both testing and benchmark purposes. Artificial datasets give the user opportunity to specify the stream properties such as noise ratio, concept drift, cluster shapes and densities. Artificial data stream generation by MOA and details of popular datasets are given. All datasets mentioned in this section, except Charitable Donation Dataset, have true class labels. Table~\ref{tbl:dataSets} summarizes properties of popular datasets.

\begin{table*}[t]
	\caption{Properties of popular datasets.}
	\centering
	\begin{tabular}{p{3cm} p{2cm} p{2cm} p{2cm}}
	%\begin{tabular}{l c c c}
		\textbf{Dataset Name} & \textbf{Number of \newline Instances} & \textbf{Number of \newline Features} & \textbf{Number of \newline Clusters} \\
		%\textbf{Dataset Name} & \textbf{Number of Instances} & \textbf{Number of Features} & \textbf{Number of Clusters} \\
		\hline
		Forest Cover Type 			& 581,012 & 54 & 7 \\
		Network Intrusion \newline Detection & 4,898,431 & 41 & 23 \\
		Network Intrusion \newline Detection Subset & 494,021 & 41 & 23 \\
		Charitable Donation			& 191,779 & 481 & \textit{not specified} \\
		Sensor Stream 	            & 2,219,803 & 5 & 54 \\
		Power Supply Stream 		& 29,928 & 2 & 24 \\
		Hyper Plane Stream 			& 100,000 & 10 & 5 \\		
		\hline
	\end{tabular}
	\label{tbl:dataSets}
\end{table*}

\subsubsection{Artificial Data Streams}
\label{subsubsec:artificalStreams}
\citeauthor{Moa} (described in Section~\ref{subsec:moa}) has a number of classes~\citep{Moa:stream:generators} to generate artificial data streams in different shapes and with or without concept drift. 

\subsubsection{Forest Cover Type Dataset}
Forest Cover Type Dataset is publicly available on Machine Learning Repository of UCI. It has totally 581,012 instances and each of them belongs to one of 7 cover types. The instances are described by 54 features, 10 of which are quantitative and 44 of which are binary. Each instance is giving information of an area of 30x30 meters. This dataset is not actually a data stream, but a stationary dataset. It does not have a time stamp or an exclusive order information. However, it is converted into a data stream by taking the data input order as the streaming order.

\subsubsection{Network Intrusion Detection Dataset}
Network Intrusion Detection Dataset is used for The Third International Knowledge Discovery and Data Mining Tools Competition, which was a session of KDD-99, The Fifth International Conference on Knowledge Discovery and Data Mining. It is publicly available on KDD archive of UCI. This set has 4,898,431 records of network traffic data and each of them belongs to one of 23 types of connection (22 attack types and normal connection). The instances are described by 41 features, some of which are discrete and the others are continuous. There exists also a 10\% subset of this dataset which is more concentrated than the original dataset. The subset itself is yet another most used dataset.

\subsubsection{Charitable Donation Dataset}
Charitable Donation Dataset is used for The Second International Knowledge Discovery and Data Mining Tools Competition, which was held in conjunction with KDD-98, The Fourth International Conference on Knowledge Discovery and Data Mining. This dataset has 191,779 instances and each instance has 481 features. These instances, are information about people who have made charitable donations in response to direct mailing requests. This dataset is publicly available on KDD archive of UCI.

\subsubsection{Various Spam Mail Datasets}
There exist several spam mail datasets publicly available in different online data repositories. Spam mail datasets are suitable for stream clustering because mails inherently are data streams. They have a date-time information which makes them easily interpreted as data streams.

\subsubsection{Various Sensor Network Datasets}
There exist several sensor network datasets publicly available on the Internet. One of sensor network data repositories is~\citeauthor{rawdad}. It is very common to use sensor network datasets in data stream clustering, since they inherently are data streams. 

\section{Data Stream Processing Tools}
\label{sec:streamTools}
We provide brief information about popular tools that are used for data stream mining.

\subsection{MOA}
\label{subsec:moa}
\citeauthor{Moa}~\citep{Bifet:MOA} is a popular open source framework for data stream mining. It is implemented in Java and released under the GNU General Public License. MOA is specialized for data streams. It includes algorithms for regression, clustering, classification, outlier detection, concept drift detection and recommender systems, and it also includes tools for evaluation. Data stream generators are provided. It can be used as both a stream processing tool and an environment to develop stream processing algorithms. Furthermore, MOA has the ability to interact with~\citet{Weka}, which is a data mining software.

\subsection{RapidMiner}
\citet{RapidMiner}, formerly known as Yet Another Learning Environment (YALE), is another data mining tool but it is developed by a private company. It has an integrated development environment, which is called RapidMiner Studio. It supports all data preparation, result visualization, model validation and optimization steps of the machine learning process. It has a \textit{Streams} plugin~\citep{RapidMiner:Streams} which integrates the stream oriented processing into the RapidMiner suite. This plugin allows developing data stream processing tools using utilities of RapidMiner.

\subsection{R}
\citet{RProject} is a free software environment and programming language for statistical computing. R is an open source project and it is released under the GNU General Public License. R, a rich in packages software environment, has special packages for clustering, data streams, stream mining etc. These packages are as follows.
\begin{itemize}
	\item stream: A framework for data stream modeling and associated data mining tasks such as clustering and classification.
	\item rstream: Unified object oriented interface for multiple independent streams of random numbers from different sources.
	\item streamMOA: Interface for data stream clustering algorithms implemented in the MOA framework.
	\item RMOA: Connects R with MOA framework to build classification and regression models on streaming data.	
\end{itemize}

\section{Data Stream Processing Platforms}
\label{sec:streamPlatforms}
Currently there exist several data stream mining platforms~\citep{platforms:JANARDAN2017804, platforms:prasad2016stream} developed by different organizations. 
\begin{itemize}
	\item \textbf{\citet{ApacheStorm}} is a distributed, real time stream processing computation framework. It is free and open source. Moreover, Apache Storm is scalable and fault tolerant. It is designed to be used with any programming language.
	\item \textbf{\citet{ApacheSpark}} is a well known, open source, fast and general engine for large-scale data processing. Apache Spark has an extension, called \textbf{\citet{SparkStreaming}}, that enables scalable, high-throughput, fault tolerant stream processing of live data streams. Spark Streaming can be seen as a layer between data streams and Apache Spark. Spark Streaming gets a data stream, creates data batches from the stream and feeds Apache Spark with these batches. In this way, results of the data stream processing are produced by Apache Spark batch by batch. Spark Stream accepts input from many different sources such as Kafka, Flume, Twitter, ZeroMQ, Kinesis, or TCP sockets.
	\item \textbf{\citet{ApacheSamza}}~\citep{SamzaLinkedin} is another open source, distributed stream processing framework. It is near real time and asynchronous. It provides fault tolerance, processor isolation, security, and resource management using Apache Hadoop Yarn. It uses Apache Kafka for messaging. Apache Samza, together with Apache Kafka, is developed by LinkedIn engineers, and commonly known as LinkedIn’s framework for stream processing.
	\item \textbf{\citet{ApacheKafka}} is an open source stream processing software platform. The objective of the project is to provide a unified, high throughput, low latency platform for real time data streams. It is scalable and fault tolerant. It has a publish-subscribe messaging system. Apache Kafka is the other platform developed by LinkedIn, similar to Apache Samza.
	\item \textbf{\citet{AmazonKinesis}} is one of the Amazon web services. It is a cloud based, real time data processing service that is developed for large and distributed data streams. In functionality, Amazon Kinesis has similarities to Apache Kafka. It is scalable and able to pull any amount of data, from any number of sources. It is designed to make it easier to develop real time applications and it has a fully managed infrastructure. 
	\item \textbf{\citet{IBMInfosphere}}~\citep{Gedik:2012:MFB:2389474.2389477} is a commercial, enterprise-grade stream processing platform, that is designed to retrieve meaningful information from data in motion, working on time window models with windows of minutes to hours. It provides low latency for time critical applications such as fraud detection and network management. It also has the ability to fuse streams. IBM Inforsphere adapts rapidly to changing data forms and types and it manages high availability itself.
	\item \textbf{\citet{GoogleStream}} is Google's solution for data stream processing. It has a fully managed infrastructure and it provides ingesting, processing and analyzing event streams in real time. It is an integrated, scalable and open stream analytics solution. Google Cloud Stream works with a full harmony with other solutions of Google Cloud, like Cloud Pub/Sub, Cloud Dataflow, BigQuery, Cloud Machine Learning etc.
	\item \textbf{\citet{Azure}} is Microsoft's solution for data stream processing. It is a serverless, scalable, on demand real time, complex event processing engine. It is able to run on multiple streams from different sources. Azure Stream Analytics has a declarative SQL like language. It can be used as integrated with other Azure solutions such as Azure Machine Learning, Azure IoT Hub, Power BI etc.
\end{itemize}

%\pagebreak

%
\section{Conclusions}
\label{sec:conclusion}
With the technological improvements, number of interconnected devices is increasing. Connected devices continuously generate large scale data with high speed, which are called data streams. Therefore, processing data streams in real time is arousing more interest and clustering seems to be the most suitable data processing method for data streams.

We present a survey of recent progress in data stream clustering algorithms. There are essential differences between traditional data clustering algorithms and data stream clustering algorithms. We emphasize the most important data stream clustering concepts such as concept drift, window models, outlier detection methods and data structures. Seven most recent data stream clustering algorithms are analyzed in detail. For each algorithm, a comprehensive analysis is presented including algorithmic detail, evaluation of the results and complexity. Global comparison of these algorithms highlighting their advantages and disadvantages is also presented.
An overview of the most popular stream processing tools and platforms is given along with stream datasets.

Several open challenges exist regarding data stream clustering. Finding number of clusters and adopting to changes in the number of clusters in data streams are the most crucial challenges. Furthermore, existing algorithms need critical parameters that directly affect clustering quality and require prior knowledge about input data. Moreover, concept drift may change data characteristics and invalidate these parameters. Developing generic and self-adapting algorithms is another popular data stream clustering challenge. Additionally, there is a lack of algorithms that handle different data types. Most of existing algorithms are able to deal with only quantitative data. Last but not least, data stream clustering algorithms should execute with high performance in despite of memory restrictions. 

It may be ideal to compare the efficiency and the effectiveness of the data stream clustering algorithms on a benchmarking framework under controlled conditions of artificial datasets that contain concept drift, outliers and class labels and of real datasets. Data stream clustering using deep neural network models and within edge computing are the two emerging topics to be explored further.

%\begin{acknowledgements}
%\end{acknowledgements}

%\pagebreak

%\section*{References}
\bibliography{mybibfile}

\begin{thebibliography}{107}
\providecommand{\natexlab}[1]{#1}
\providecommand{\url}[1]{{#1}}
\providecommand{\urlprefix}{URL }
\expandafter\ifx\csname urlstyle\endcsname\relax
  \providecommand{\doi}[1]{DOI~\discretionary{}{}{}#1}\else
  \providecommand{\doi}{DOI~\discretionary{}{}{}\begingroup
  \urlstyle{rm}\Url}\fi
\providecommand{\eprint}[2][]{\url{#2}}

\bibitem[{{A Community Resource for Archiving Wireless Data At Dartmouth
  (CRAWDAD)}(n.d.)}]{rawdad}
{A Community Resource for Archiving Wireless Data At Dartmouth (CRAWDAD)}
  (n.d.) \url{https://crawdad.org/keyword-sensor-network.html}, accessed:
  2018-03-25

\bibitem[{Ackermann et~al.(2012)Ackermann, M\"{a}rtens, Raupach, Swierkot,
  Lammersen, and Sohler}]{Ackermann:2012:streamKM++}
Ackermann MR, M\"{a}rtens M, Raupach C, Swierkot K, Lammersen C, Sohler C
  (2012) Streamkm++: A clustering algorithm for data streams. J Exp
  Algorithmics 17:2.4:2.1--2.4:2.30

\bibitem[{Aggarwal et~al.(2004)Aggarwal, Han, Wang, and Yu}]{2004:hpstream}
Aggarwal C, Han J, Wang J, Yu P (2004) A framework for projected clustering of
  high dimensional data streams. pp 852--863,
  \doi{10.1016/B978-012088469-8/50075-9}

\bibitem[{Aggarwal(2013)}]{Aggarwal2013chapter}
Aggarwal CC (2013) A survey of stream clustering algorithms. In: Aggarwal CC,
  Reddy CK (eds) Data Clustering: Algorithms and Applications, {CRC} Press, pp
  231--258

\bibitem[{Aggarwal et~al.(2003)Aggarwal, Han, Wang, and
  Yu}]{Aggarwal:2003:Clustream}
Aggarwal CC, Han J, Wang J, Yu PS (2003) A framework for clustering evolving
  data streams. In: Proceedings of the 29th International Conference on Very
  Large Data Bases - Volume 29, VLDB '03, pp 81--92

\bibitem[{Agrawal et~al.(1998)Agrawal, Gehrke, Gunopulos, and
  Raghavan}]{1998:clique}
Agrawal R, Gehrke J, Gunopulos D, Raghavan P (1998) Automatic subspace
  clustering of high dimensional data for data mining applications. In:
  Proceedings of the 1998 ACM SIGMOD International Conference on Management of
  Data, Association for Computing Machinery, New York, NY, USA, SIGMOD ’98, p
  94–105, \doi{10.1145/276304.276314},
  \urlprefix\url{https://doi.org/10.1145/276304.276314}

\bibitem[{Alam et~al.(2016)Alam, Mehmood, Katib, and
  Albeshri}]{ALAM:2016:review}
Alam F, Mehmood R, Katib I, Albeshri A (2016) Analysis of eight data mining
  algorithms for smarter internet of things (iot). Procedia Computer Science
  98:437 -- 442

\bibitem[{AmazonKinesis(2013)}]{AmazonKinesis}
AmazonKinesis (2013) {Amazon Kinesis}. \url{https://aws.amazon.com/kinesis/},
  accessed: 2018-03-25

\bibitem[{Amini et~al.(2016)Amini, Saboohi, Herawan, and
  Wah}]{Amini:2016:MudiStream}
Amini A, Saboohi H, Herawan T, Wah TY (2016) Mudi-stream: A multi density
  clustering algorithm for evolving data stream. J Netw Comput Appl
  59(C):370--385

\bibitem[{Andrade~Silva et~al.(2017)Andrade~Silva, Hruschka, and
  Gama}]{Silva:2017:FEAC-Stream}
Andrade~Silva Jd, Hruschka ER, Gama Ja (2017) An evolutionary algorithm for
  clustering data streams with a variable number of clusters. Expert Syst Appl
  67(C):228--238

\bibitem[{{Apache Kafka}(2011)}]{ApacheKafka}
{Apache Kafka} (2011) \url{https://kafka.apache.org/}, accessed: 2018-03-25

\bibitem[{{Apache Samza}(2013)}]{ApacheSamza}
{Apache Samza} (2013) {Samza}. \url{https://samza.apache.org/}, accessed:
  2018-03-25

\bibitem[{{Apache Spark}(2012)}]{ApacheSpark}
{Apache Spark} (2012) {Apache Spark} lightning-fast cluster computing.
  \url{https://spark.apache.org/}, accessed: 2018-03-25

\bibitem[{{Apache Storm}(2011)}]{ApacheStorm}
{Apache Storm} (2011) \url{http://storm.apache.org/}, accessed: 2018-03-25

\bibitem[{Bezerra et~al.(2020)Bezerra, Costa, Guedes, and
  Angelov}]{BEZERRA:2020}
Bezerra CG, Costa BSJ, Guedes LA, Angelov PP (2020) An evolving approach to
  data streams clustering based on typicality and eccentricity data analytics.
  Information Sciences 518:13 -- 28,
  \doi{https://doi.org/10.1016/j.ins.2019.12.022},
  \urlprefix\url{http://www.sciencedirect.com/science/article/pii/S0020025519311363}

\bibitem[{Bhosale(2014)}]{Bhosale2014outlier}
Bhosale SV (2014) {A Survey: Outlier Detection in Streaming Data Using
  Clustering Approached}. (IJCSIT) International Journal of Computer Science
  and Information Technologies 5:6050--6053

\bibitem[{Bifet et~al.(2010)Bifet, Holmes, Kirkby, and Pfahringer}]{Bifet:MOA}
Bifet A, Holmes G, Kirkby R, Pfahringer B (2010) Moa: Massive online analysis.
  J Mach Learn Res 11:1601--1604,
  \urlprefix\url{http://dl.acm.org/citation.cfm?id=1756006.1859903}

\bibitem[{Bockermann(2018)}]{RapidMiner:Streams}
Bockermann C (2018) {RapidMiner Streams Plugin}.
  \url{https://sfb876.de/streams/doc/rapidminer.html}, accessed: 2018-03-25

\bibitem[{Cao et~al.(2006)Cao, Ester, Qian, and Zhou}]{2006:denstream}
Cao F, Ester M, Qian W, Zhou A (2006) Density-based clustering over an evolving
  data stream with noise. vol 2006, \doi{10.1137/1.9781611972764.29}

\bibitem[{Carnein et~al.(2017)Carnein, Assenmacher, and
  Trautmann}]{Carnein:2017:review}
Carnein M, Assenmacher D, Trautmann H (2017) An empirical comparison of stream
  clustering algorithms. In: Proceedings of the Computing Frontiers Conference,
  CF'17, pp 361--366

\bibitem[{Chauhan and Shukla(2015)}]{Chauhan2015outlier}
Chauhan P, Shukla M (2015) {A review on outlier detection techniques on data
  stream by using different approaches of K-Means algorithm}. In: 2015
  International Conference on Advances in Computer Engineering and Applications

\bibitem[{Chen and Tu(2007)}]{Chen:2007:D-Stream}
Chen Y, Tu L (2007) Density-based clustering for real-time stream data. In:
  Proceedings of the 13th ACM SIGKDD International Conference on Knowledge
  Discovery and Data Mining, KDD '07, pp 133--142

\bibitem[{{Christodoulou} et~al.(2018){Christodoulou}, {Bi}, and
  {Wilkie}}]{Christodoulou2018timeseries}
{Christodoulou} V, {Bi} Y, {Wilkie} G (2018) A fuzzy shape-based anomaly
  detection and its application to electromagnetic data. IEEE Journal of
  Selected Topics in Applied Earth Observations and Remote Sensing
  11(9):3366--3379, \doi{10.1109/JSTARS.2018.2854865}

\bibitem[{{Citi Bike NYC}(2013)}]{CitiBike}
{Citi Bike NYC} (2013) {Citi Bike: NYC's Official Bike Sharing System}.
  \url{https://www.citibikenyc.com/}, accessed: 2018-03-25

\bibitem[{{Citi Bike System Data}(2013)}]{CitiBike:SystemData}
{Citi Bike System Data} (2013) \url{https://www.citibikenyc.com/system-data},
  accessed: 2018-03-25

\bibitem[{Dang et~al.(2009)Dang, Lee, Ng, and Ong}]{2009:swem}
Dang XH, Lee VCS, Ng WK, Ong KL (2009) Incremental and adaptive clustering
  stream data over sliding window. In: Bhowmick SS, K{\"u}ng J, Wagner R (eds)
  Database and Expert Systems Applications, Springer Berlin Heidelberg, Berlin,
  Heidelberg, pp 660--674

\bibitem[{Din et~al.(2020)Din, Shao, Kumar, Ali, Liu, and Ye}]{2020:UDDIN}
Din SU, Shao J, Kumar J, Ali W, Liu J, Ye Y (2020) Online reliable
  semi-supervised learning on evolving data streams. Information Sciences
  525:153 -- 171, \doi{https://doi.org/10.1016/j.ins.2020.03.052},
  \urlprefix\url{http://www.sciencedirect.com/science/article/pii/S0020025520302322}

\bibitem[{Ding et~al.(2015)Ding, Wu, Qian, Jia, and Jin}]{Ding:2015:review}
Ding S, Wu F, Qian J, Jia H, Jin F (2015) Research on data stream clustering
  algorithms. Artif Intell Rev 43(4):593--600

\bibitem[{Duan et~al.(2006)Duan, Xiong, Lee, and Guo}]{2006:ldbscan}
Duan L, Xiong D, Lee J, Guo F (2006) A local density based spatial clustering
  algorithm with noise. vol~32, pp 4061--4066, \doi{10.1109/ICSMC.2006.384769}

\bibitem[{Ester et~al.(1998)Ester, Kriegel, Sander, Wimmer, and
  Xu}]{1998:incrementalDBSCAN}
Ester M, Kriegel HP, Sander J, Wimmer M, Xu X (1998) Incremental clustering for
  mining in a data warehousing environment. In: Proceedings of the 24rd
  International Conference on Very Large Data Bases, Morgan Kaufmann Publishers
  Inc., San Francisco, CA, USA, VLDB ’98, p 323–333

\bibitem[{{Fahy} et~al.(2018){Fahy}, {Yang}, and
  {Gongora}}]{Fahy:2018:AntColony}
{Fahy} C, {Yang} S, {Gongora} M (2018) Ant colony stream clustering: A fast
  density clustering algorithm for dynamic data streams. IEEE Transactions on
  Cybernetics pp 1--14

\bibitem[{Fisher(1996)}]{1996:cobweb}
Fisher D (1996) Iterative optimization and simplification of hierarchical
  clustering. J Artif Intell Res 4, \doi{10.1613/jair.276}

\bibitem[{Gaber et~al.(2009)Gaber, Zaslavsky, and Krishnaswamy}]{gaber2009data}
Gaber MM, Zaslavsky A, Krishnaswamy S (2009) Data stream mining. In: Data
  Mining and Knowledge Discovery Handbook, Springer, pp 759--787

\bibitem[{Gama et~al.(2011)Gama, Rodrigues, and Lopes}]{Gama:2011:DGClust}
Gama Ja, Rodrigues PP, Lopes L (2011) Clustering distributed sensor data
  streams using local processing and reduced communication. Intell Data Anal
  15(1):3--28

\bibitem[{Gama et~al.(2014)Gama, \v{Z}liobaite, Bifet, Pechenizkiy, and
  Bouchachia}]{conceptDriftSurvey}
Gama Ja, \v{Z}liobaite I, Bifet A, Pechenizkiy M, Bouchachia A (2014) A survey
  on concept drift adaptation. ACM Comput Surv 46(4):44:1--44:37

\bibitem[{Gedik and Andrade(2012)}]{Gedik:2012:MFB:2389474.2389477}
Gedik B, Andrade H (2012) A model-based framework for building extensible, high
  performance stream processing middleware and programming language for ibm
  infosphere streams. Softw Pract Exper 42(11):1363--1391

\bibitem[{Ghesmoune et~al.(2016)Ghesmoune, Lebbah, and
  Azzag}]{Ghesmoune:2016:review}
Ghesmoune M, Lebbah M, Azzag H (2016) State-of-the-art on clustering data
  streams. Big Data Analytics 1(1):13

\bibitem[{{Google Cloud Stream}(2012)}]{GoogleStream}
{Google Cloud Stream} (2012) {Streaming Analytics for Real Time Insights -
  Google Cloud}.
  \url{https://cloud.google.com/solutions/big-data/stream-analytics/},
  accessed: 2018-03-25

\bibitem[{Hassani et~al.(2014)Hassani, Spaus, and
  Seidl}]{Hassani:2014:HASTREAM}
Hassani M, Spaus P, Seidl T (2014) Adaptive multiple-resolution stream
  clustering. In: Machine Learning and Data Mining in Pattern Recognition, pp
  134--148

\bibitem[{Hassani et~al.(2015)Hassani, Spaus, Cuzzocrea, and
  Seidl}]{Hassani:2015:I-HASTREAM}
Hassani M, Spaus P, Cuzzocrea A, Seidl T (2015) Adaptive stream clustering
  using incremental graph maintenance. In: Proceedings of the 4th International
  Conference on Big Data, Streams and Heterogeneous Source Mining: Algorithms,
  Systems, Programming Models and Applications - Volume 41, BIGMINE'15, pp
  49--64

\bibitem[{Hassani et~al.(2016)Hassani, Spaus, Cuzzocrea, and
  Seidl}]{Hassani:2016:I-HASTREAM}
Hassani M, Spaus P, Cuzzocrea A, Seidl T (2016) I-hastream: Density-based
  hierarchical clustering of big data streams and its application to big graph
  analytics tools. In: 2016 16th IEEE/ACM International Symposium on Cluster,
  Cloud and Grid Computing (CCGrid), pp 656--665

\bibitem[{Hyde et~al.(2017)Hyde, Angelov, and MacKenzie}]{Hyde:2017:Cedas}
Hyde R, Angelov P, MacKenzie A (2017) Fully online clustering of evolving data
  streams into arbitrarily shaped clusters. Information Sciences 382-383:96 --
  114

\bibitem[{{IBM Infosphere}(1996)}]{IBMInfosphere}
{IBM Infosphere} (1996) {Streaming Analytics - Overview - IBM Cloud}.
  \url{https://www.ibm.com/cloud/streaming-analytics}, accessed: 2018-03-25

\bibitem[{Isaksson et~al.(2012)Isaksson, Dunham, and Hahsler}]{2012:sostream}
Isaksson C, Dunham M, Hahsler M (2012) Sostream: Self organizing density-based
  clustering over data stream. vol 7376, \doi{10.1007/978-3-642-31537-4_21}

\bibitem[{Janardan and Mehta(2017)}]{platforms:JANARDAN2017804}
Janardan, Mehta S (2017) Concept drift in streaming data classification:
  Algorithms, platforms and issues. Procedia Computer Science 122:804 -- 811,
  \doi{https://doi.org/10.1016/j.procs.2017.11.440},
  \urlprefix\url{http://www.sciencedirect.com/science/article/pii/S1877050917326881},
  5th International Conference on Information Technology and Quantitative
  Management, ITQM 2017

\bibitem[{Karypis et~al.(1999)Karypis, Han, and Kumar}]{Karypis:1999:CHAMELEON}
Karypis G, Han EH, Kumar V (1999) Chameleon a hierarchical clustering algorithm
  using dynamic modeling. Computer 32:68 -- 75, \doi{10.1109/2.781637}

\bibitem[{Kaufman and Rousseeuw(1990)}]{2008:clara}
Kaufman L, Rousseeuw PJ (1990) Clustering Large Applications (Program CLARA),
  John Wiley \& Sons, Ltd, chap~3, pp 126--163.
  \doi{10.1002/9780470316801.ch3},
  \urlprefix\url{https://onlinelibrary.wiley.com/doi/abs/10.1002/9780470316801.ch3},
  \eprint{https://onlinelibrary.wiley.com/doi/pdf/10.1002/9780470316801.ch3}

\bibitem[{Keogh et~al.(2005)Keogh, Lin, and Fu}]{Keogh2005timeseries}
Keogh E, Lin J, Fu A (2005) Hot sax: Efficiently finding the most unusual time
  series subsequence. In: Proceedings of the Fifth IEEE International
  Conference on Data Mining, IEEE Computer Society, USA, ICDM ’05, p
  226–233, \doi{10.1109/ICDM.2005.79},
  \urlprefix\url{https://doi.org/10.1109/ICDM.2005.79}

\bibitem[{Kim and Park(2020)}]{KIM:2020}
Kim T, Park CH (2020) Anomaly pattern detection for streaming data. Expert
  Systems with Applications 149:113252,
  \doi{https://doi.org/10.1016/j.eswa.2020.113252},
  \urlprefix\url{http://www.sciencedirect.com/science/article/pii/S0957417420300774}

\bibitem[{Kong et~al.(2019)Kong, Bi, and Glass}]{Kong2019timeseries}
Kong X, Bi Y, Glass DH (2019) Detecting anomalies in sequential data augmented
  with new features. Artificial Intelligence Review 53:625--652

\bibitem[{Kremer et~al.(2011)Kremer, Kranen, Jansen, Seidl, Bifet, Holmes, and
  Pfahringer}]{Kremer:2011:CMM}
Kremer H, Kranen P, Jansen T, Seidl T, Bifet A, Holmes G, Pfahringer B (2011)
  An effecive evaluation measure for clustering on evolving data streams. In:
  Proceedings of the 17th ACM SIGKDD International Conference on Knowledge
  Discovery and Data Mining, KDD '11, pp 868--876

\bibitem[{Kumar(2016)}]{Kumar:2016:review}
Kumar P (2016) {Data Stream Clustering in Internet of Things}. SSRG
  International Journal of Computer Science and Engineering 3(8)

\bibitem[{{Liu} et~al.(2009){Liu}, {Huang}, {Guo}, and
  {Chen}}]{2009:rDenStream}
{Liu} L, {Huang} H, {Guo} Y, {Chen} F (2009) rdenstream, a clustering algorithm
  over an evolving data stream. In: 2009 International Conference on
  Information Engineering and Computer Science, pp 1--4

\bibitem[{Lu et~al.(2005)Lu, Sun, Xu, and Liu}]{2005:gchds}
Lu Y, Sun Y, Xu G, Liu G (2005) A grid-based clustering algorithm for
  high-dimensional data streams. In: Li X, Wang S, Dong ZY (eds) Advanced Data
  Mining and Applications, Springer Berlin Heidelberg, Berlin, Heidelberg, pp
  824--831

\bibitem[{Mahdiraji(2009)}]{Mahdiraji2009survey}
Mahdiraji AR (2009) Clustering data stream: A survey of algorithms. Int J
  Know-Based Intell Eng Syst 13(2):39–44

\bibitem[{Mansalis et~al.(2018)Mansalis, Ntoutsi, Pelekis, and
  Theodoridis}]{Mansalis:2018:review}
Mansalis S, Ntoutsi E, Pelekis N, Theodoridis Y (2018) An evaluation of data
  stream clustering algorithms. Statistical Analysis and Data Mining: The ASA
  Data Science Journal 11(4):167--187

\bibitem[{{Massive Online Analysis (MOA)}(2014)}]{Moa}
{Massive Online Analysis (MOA)} (2014) {Moa - Machine Learning for Data
  Streams}. \url{https://moa.cms.waikato.ac.nz/}, accessed: 2018-03-25

\bibitem[{Meesuksabai et~al.(2011)Meesuksabai, Kangkachit, and
  Waiyamai}]{2011:HUE-Stream}
Meesuksabai W, Kangkachit T, Waiyamai K (2011) Hue-stream: Evolution-based
  clustering technique for heterogeneous data streams with uncertainty. pp
  27--40, \doi{10.1007/978-3-642-25856-5_3}

\bibitem[{Meetup(2002)}]{Meetup}
Meetup (2002) {We are what we do | Meetup}. \url{https://www.meetup.com/},
  accessed: 2018-03-25

\bibitem[{{Meetup Stream}(2002)}]{Meetup:RSVP}
{Meetup Stream} (2002) {Extend your community | Meetup}.
  \url{https://www.meetup.com/meetup_api/docs/stream/2/rsvps/}, accessed:
  2018-03-25

\bibitem[{Merino(2015)}]{Merino2015}
Merino JA (2015) {Streaming Data Clustering in MOA using the Leader Algorithm}.
  PhD thesis, Universitat Polit\`{e}cnica de Catalunya

\bibitem[{{Microsoft Azure Stream Analytics}(2012)}]{Azure}
{Microsoft Azure Stream Analytics} (2012) {Stream Analytics - Real Time Data
  Analytics - Microsoft Azure}.
  \url{https://azure.microsoft.com/en-us/services/stream-analytics/}, accessed:
  2018-03-25

\bibitem[{{Moa Stream Generators}(2014)}]{Moa:stream:generators}
{Moa Stream Generators} (2014) {Moa: Package moa.stream.generators}.
  \url{https://www.cs.waikato.ac.nz/~abifet/MOA/API/namespacemoa_1_1streams_1_1generators.html},
  accessed: 2018-03-25

\bibitem[{Modi and Oza(2017)}]{Modi2017OutlierAA}
Modi KD, Oza PB (2017) Outlier analysis approaches in data mining.
  International Journal of Innovative Research in Technology (IJIRT) 3:6--12

\bibitem[{Mousavi et~al.(2015)Mousavi, Bakar, and
  Vakilian}]{Mousavi:2015:review}
Mousavi M, Bakar A, Vakilian M (2015) Data stream clustering algorithms: A
  review. International Journal of Advances in Soft Computing and its
  Applications 7:1--15

\bibitem[{Mouss et~al.(2004)Mouss, Mouss, Mouss, and Sefouhi}]{Page-Hinckley}
Mouss H, Mouss D, Mouss N, Sefouhi L (2004) Test of page-hinckley, an approach
  for fault detection in an agro-alimentary production system. In: 2004 5th
  Asian Control Conference (IEEE Cat. No.04EX904), vol~2, pp 815--818 Vol.2

\bibitem[{Namadchian and Esfandani(2012)}]{Namadchian2012DSCLUAN}
Namadchian A, Esfandani G (2012) Dsclu: A new data stream clustring algorithm
  for multi density environments. 2012 13th ACIS International Conference on
  Software Engineering, Artificial Intelligence, Networking and
  Parallel/Distributed Computing pp 83--88

\bibitem[{{National Weather Service (NWS)}(1870)}]{NWS}
{National Weather Service (NWS)} (1870) {National Weather Service}.
  \url{https://www.weather.gov/}, accessed: 2018-03-25

\bibitem[{Nguyen et~al.(2015)Nguyen, Woon, and Ng}]{Nguyen:2015:review}
Nguyen HL, Woon YK, Ng WK (2015) A survey on data stream clustering and
  classification. Knowledge and Information Systems 45(3):535--569

\bibitem[{{NWS Public Alerts}(n.d.)}]{NWS:alerts}
{NWS Public Alerts} (n.d.) {NWS Public Alerts}.
  \url{https://alerts.weather.gov/}, accessed: 2018-03-25

\bibitem[{O'Callaghan et~al.(2002)O'Callaghan, Meyerson, Motwani, Mishra, and
  Guha}]{Callagha:2002:LSearch}
O'Callaghan L, Meyerson A, Motwani R, Mishra N, Guha S (2002) Streaming-data
  algorithms for high-quality clustering. In: Proceedings of the 18th
  International Conference on Data Engineering, ICDE '02, pp 685--

\bibitem[{Ordonez(2003)}]{2003:incrementalkmeans}
Ordonez C (2003) Clustering binary data streams with k-means. In: Proceedings
  of the 8th ACM SIGMOD Workshop on Research Issues in Data Mining and
  Knowledge Discovery, Association for Computing Machinery, New York, NY, USA,
  DMKD ’03, p 12–19, \doi{10.1145/882082.882087},
  \urlprefix\url{https://doi.org/10.1145/882082.882087}

\bibitem[{Parzen(1962)}]{parzen:1962:kde}
Parzen E (1962) On estimation of a probability density function and mode. Ann
  Math Statist 33(3):1065--1076

\bibitem[{Prasad and Agarwal(2016)}]{platforms:prasad2016stream}
Prasad BR, Agarwal S (2016) Stream data mining: platforms, algorithms,
  performance evaluators and research trends. International journal of database
  theory and application 9(9):201--218

\bibitem[{Puschmann et~al.(2017)Puschmann, Barnaghi, and
  Tafazolli}]{Puschmann:2017:adaptiveStreaming}
Puschmann D, Barnaghi P, Tafazolli R (2017) Adaptive clustering for dynamic iot
  data streams. IEEE Internet of Things Journal 4(1):64--74

\bibitem[{R(1993)}]{RProject}
R (1993) {R - The R Project for Statistical Computing}.
  \url{https://www.r-project.org/}, accessed: 2018-03-25

\bibitem[{Ramesh(2013)}]{SamzaLinkedin}
Ramesh N (2013) {Apache Samza, LinkedIn's Framework for Stream Processing - The
  New Stack}.
  \url{https://thenewstack.io/apache-samza-linkedins-framework-for-stream-processing/},
  accessed: 2018-03-25

\bibitem[{Ramirez-Gallego et~al.(2017)Ramirez-Gallego, Krawczyk, Garcia,
  Wozniak, and Herrera}]{RamirezDataPreprocessing}
Ramirez-Gallego S, Krawczyk B, Garcia S, Wozniak M, Herrera F (2017) A survey
  on data preprocessing for data stream mining: Current status and future
  directions. Neurocomputing 239

\bibitem[{{RapidMiner}(2001)}]{RapidMiner}
{RapidMiner} (2001) {Data Sicence Platform - RapidMiner}.
  \url{https://rapidminer.com/}, accessed: 2018-03-25

\bibitem[{Rodrigues et~al.(2006)Rodrigues, Gama, and
  Pedroso}]{Rodrigues:2006:ODAC}
Rodrigues P, Gama J, Pedroso JP (2006) Odac: Hierarchical clustering of time
  series data streams. \doi{10.1137/1.9781611972764.48}

\bibitem[{Rosenblatt(1956)}]{rosenblatt:1956:kde}
Rosenblatt M (1956) Remarks on some nonparametric estimates of a density
  function. Ann Math Statist 27(3):832--837

\bibitem[{Sadik and Gruenwald(2014)}]{Sadik:2014outlier}
Sadik S, Gruenwald L (2014) Research issues in outlier detection for data
  streams. SIGKDD Explor Newsl 15(1):33--40

\bibitem[{{Satyanarayanan}(2017)}]{Satyanarayanan2017edge}
{Satyanarayanan} M (2017) The emergence of edge computing. Computer
  50(1):30--39, \doi{10.1109/MC.2017.9}

\bibitem[{Sheikholeslami et~al.(2000)Sheikholeslami, Chatterjee, and
  Zhang}]{2000:WaveCluster}
Sheikholeslami G, Chatterjee S, Zhang A (2000) Wavecluster: A wavelet-based
  clustering approach for spatial data in very large databases. The VLDB
  Journal 8(3–4):289–304, \doi{10.1007/s007780050009},
  \urlprefix\url{https://doi.org/10.1007/s007780050009}

\bibitem[{Shi and Dustdar(2016)}]{ShiDustdar:2016:Edge}
Shi W, Dustdar S (2016) The promise of edge computing. Computer 49(5):78--81

\bibitem[{Shi et~al.(2016)Shi, Cao, Zhang, Li, and Xu}]{ShiCao:2016:Edge}
Shi W, Cao J, Zhang Q, Li Y, Xu L (2016) Edge computing: Vision and challenges.
  IEEE Internet of Things Journal 3(5):637--646

\bibitem[{Silva et~al.(2013)Silva, Faria, Barros, Hruschka, Carvalho, and
  Gama}]{Silva:2013survey}
Silva JA, Faria ER, Barros RC, Hruschka ER, Carvalho ACPLFd, Gama Ja (2013)
  Data stream clustering: A survey. ACM Comput Surv 46(1):13:1--13:31

\bibitem[{Song and Kasabov(2001)}]{Qun:2001:Ecm}
Song Q, Kasabov N (2001) Ecm - a novel on-line, evolving clustering method and
  its applications. In: In M. I. Posner (Ed.), Foundations of cognitive
  science, The MIT Press, pp 631--682

\bibitem[{Souiden et~al.(2016)Souiden, Brahmi, and Toumi}]{Souiden2016outlier}
Souiden I, Brahmi Z, Toumi H (2016) {A Survey On Outlier Detection In The
  Context Of Stream Mining : review Of Existing Approaches And Recommadations}.
  In: Advances in Intelligent Systems and Computing

\bibitem[{{Spark Streaming}(2012)}]{SparkStreaming}
{Spark Streaming} (2012) {Apache Spark Streaming}.
  \url{https://spark.apache.org/streaming/}, accessed: 2018-03-25

\bibitem[{Sun and Lu(2006)}]{2006:gscds}
Sun Y, Lu Y (2006) A grid-based subspace clustering algorithm for
  high-dimensional data streams. In: Feng L, Wang G, Zeng C, Huang R (eds) Web
  Information Systems -- WISE 2006 Workshops, Springer Berlin Heidelberg,
  Berlin, Heidelberg, pp 37--48

\bibitem[{Tasoulis et~al.(2007)Tasoulis, Ross, and Adams}]{2007:opticsstream}
Tasoulis D, Ross G, Adams N (2007) Visualising the cluster structure of data
  streams. vol 4723, pp 81--92, \doi{10.1007/978-3-540-74825-0_8}

\bibitem[{Thakkar et~al.(2016)Thakkar, Vala, and
  Prajapati}]{Thakkar2016outlier}
Thakkar P, Vala J, Prajapati V (2016) {Survey on Outlier Detection in Data
  Stream}. International Journal of Computer Applications 136(2)

\bibitem[{Udommanetanakit et~al.(2007)Udommanetanakit, Rakthanmanon, and
  Waiyamai}]{2007:E-Stream}
Udommanetanakit K, Rakthanmanon T, Waiyamai K (2007) E-stream: Evolution-based
  technique for stream clustering. vol 4632, pp 605--615,
  \doi{10.1007/978-3-540-73871-8_58}

\bibitem[{{Waikato Environment for Knowledge Analysis}(1993)}]{Weka}
{Waikato Environment for Knowledge Analysis} (1993) {Weka 3 - Data Mining With
  Open Source Machine Learning Software in Java}.
  \url{https://www.cs.waikato.ac.nz/ml/weka/}, accessed: 2018-03-25

\bibitem[{Wan et~al.(2009)Wan, Ng, Dang, Yu, and Zhang}]{2009:mr-stream}
Wan L, Ng WK, Dang XH, Yu PS, Zhang K (2009) Density-based clustering of data
  streams at multiple resolutions. ACM Trans Knowl Discov Data 3(3),
  \doi{10.1145/1552303.1552307},
  \urlprefix\url{https://doi.org/10.1145/1552303.1552307}

\bibitem[{Wang et~al.(2012)Wang, Yu, Wang, and Wan}]{2012:opclustream}
Wang H, Yu Y, Wang Q, Wan Y (2012) A density-based clustering structure mining
  algorithm for data streams. In: Proceedings of the 1st International Workshop
  on Big Data, Streams and Heterogeneous Source Mining: Algorithms, Systems,
  Programming Models and Applications, Association for Computing Machinery, New
  York, NY, USA, BigMine ’12, p 69–76, \doi{10.1145/2351316.2351326},
  \urlprefix\url{https://doi.org/10.1145/2351316.2351326}

\bibitem[{Wang et~al.(1997)Wang, Yang, and Muntz}]{1997:STING}
Wang W, Yang J, Muntz RR (1997) Sting: A statistical information grid approach
  to spatial data mining. In: Proceedings of the 23rd International Conference
  on Very Large Data Bases, Morgan Kaufmann Publishers Inc., San Francisco, CA,
  USA, VLDB ’97, p 186–195

\bibitem[{Xu et~al.(2017)Xu, Wang, Li, Deng, and Gou}]{DPClust}
Xu J, Wang G, Li T, Deng W, Gou G (2017) Fat node leading tree for data stream
  clustering with density peaks. Knowledge-Based Systems 120:99 -- 117,
  \doi{https://doi.org/10.1016/j.knosys.2016.12.025},
  \urlprefix\url{http://www.sciencedirect.com/science/article/pii/S0950705116305305}

\bibitem[{Yasumoto et~al.(2016)Yasumoto, Yamaguchi, and
  Shigeno}]{Keiichi:2016:review}
Yasumoto K, Yamaguchi H, Shigeno H (2016) Survey of real-time processing
  technologies of iot data streams. Journal of Information Processing
  24(2):195--202

\bibitem[{Yin et~al.(2017)Yin, Xia, Zhang, Sun, and
  Wang}]{Yin2017improvedClustering}
Yin C, Xia L, Zhang S, Sun R, Wang J (2017) Improved clustering algorithm based
  on high-speed network data stream. Soft Computing

\bibitem[{Zhang et~al.(2017)Zhang, Zhong, Tian, Zhang, and
  Li}]{Zhang:2017:dbiecm}
Zhang KS, Zhong L, Tian L, Zhang XY, Li L (2017) {DBIECM-an Evolving Clustering
  Method for Streaming Data Clustering}. Amse Journals-Amse Iieta
  60(1):239--254

\bibitem[{Zhang et~al.(1996)Zhang, Ramakrishnan, and Livny}]{Zhang:1996:Birch}
Zhang T, Ramakrishnan R, Livny M (1996) Birch: An efficient data clustering
  method for very large databases. SIGMOD Rec 25(2):103--114

\bibitem[{{Zhang} et~al.(2014){Zhang}, {Furtlehner}, {Germain-Renaud}, and
  {Sebag}}]{2014:strAP}
{Zhang} X, {Furtlehner} C, {Germain-Renaud} C, {Sebag} M (2014) Data stream
  clustering with affinity propagation. IEEE Transactions on Knowledge and Data
  Engineering 26(7):1644--1656

\bibitem[{{Zhou} et~al.(2007){Zhou}, {Cao}, {Yan}, {Sha}, and
  {He}}]{2007:cludistream}
{Zhou} A, {Cao} F, {Yan} Y, {Sha} C, {He} X (2007) Distributed data stream
  clustering: A fast em-based approach. In: 2007 IEEE 23rd International
  Conference on Data Engineering, pp 736--745

\bibitem[{Zhou et~al.(2008)Zhou, Cao, Qian, and Jin}]{Zhou:2008:swclustering}
Zhou A, Cao F, Qian W, Jin C (2008) Tracking clusters in evolving data streams
  over sliding windows. Knowl Inf Syst 15(2):181--214

\bibitem[{Zhu(2010)}]{StreamDataRepo}
Zhu XH (2010) {Stream Data Mining Repository}.
  \url{http://www.cse.fau.edu/~xqzhu/stream.html}, accessed: 2018-03-25

\end{thebibliography}

\end{document}